\documentclass[paper,twocolumn,twoside]{geophysics}
\usepackage{amsmath}
\usepackage{enumitem}
\usepackage[normalem]{ulem}
\usepackage{amssymb}
\usepackage{xcolor}
\usepackage{algorithm}
\usepackage{algorithmic}
\usepackage{multirow}
\usepackage{booktabs}
\usepackage{graphicx}
\usepackage{bbm}
\usepackage{hyperref}
\usepackage{nameref}
\usepackage{natbib}

\begin{document}

\title{A unified framework for evaluating the robustness of machine-learning interpretability for prospect risking}
\title{Boosting Trust in Local Explanation Methods through Necessity and Sufficiency Quantification for High Dimensional Tabular Data}
\title{An Analysis of Local Interpretation Techniques for High Dimensional Tabular Data}
\title{A unified framework for evaluating robustness of Machine Learning Interpretability for Prospect Risking}
\renewcommand{\thefootnote}{\fnsymbol{footnote}} 

\address{
\footnotemark[1]Center for Energy and Geo Processing (CeGP), Omni Lab for Intelligent Visual Engineering and Science (OLIVES), School of Electrical and Computer Engineering, Georgia Institute of Technology, Atlanta, GA.}
\author{Prithwijit Chowdhury\footnotemark[1],  Ahmad Mustafa\footnotemark[1], Mohit Prabhushankar\footnotemark[1] and Ghassan AlRegib\footnotemark[1]}

\footer{Example}
\lefthead{Chowdhury et al.}
\righthead{Counterfactual Explanations}



\onecolumn 

A revised version of this manuscript has been accepted to Geophysics. The copyrights for the accepted manuscript belong strictly to the Society for Exploration Geophysicists (SEG). This document may strictly be used only for educational and other non-commercial purposes only. The full citation to the accepted
manuscript and the DOI has been listed below.

\begin{description}[labelindent=0cm,leftmargin=3cm,style=multiline]

\item[\textbf{Citation}]{P. Chowdhury, A. Mustafa, M. Prabhushankar, and G. AlRegib, "A unified framework for evaluating the robustness of machine-learning interpretability for prospect risking", accepted at  Geophysics}

\item[\textbf{DIO}]{\url{https://doi.org/10.1190/geo2024-0020.1}}

\item[\textbf{Review}]{Submission: 21 Feb 2024 \\
                        Acceptance: 07 April 2025  (2 Major \& 2 Minor revision)}

\item[\textbf{Code}]{\url{https://github.com/olivesgatech/Necessity-Sufficiency}}

\item[\textbf{Dataset}]{Data is proprietary and can not be shared.}

\item[\textbf{Bib}]{@article\{chowdhury2025unified,\\
  title=\{A unified framework for evaluating the robustness of machine-learning interpretability for prospect risking\},\\
  author=\{Chowdhury, Prithwijit and Mustafa, Ahmad and Prabhushankar, Mohit and AlRegib, Ghassan\},\\
  journal=\{Geophysics\},\\
  volume=\{90\},\\
  number=\{3\},\\
  pages=\{IM103--IM118\},\\
  year=\{2025\},\\
  publisher=\{Society of Exploration Geophysicists\}
\\
\}
}


\item[\textbf{Copyright}] {\textcopyright arXiv.org perpetual, non-exclusive license }

\item[\textbf{Contact}]{\{pchowdhury6, alregib\}@gatech.edu  \\ \url{https://alregib.ece.gatech.edu/} }

\item[\textbf{Corresponding author}]{alregib@gatech.edu }

\end{description}

\thispagestyle{empty}
\newpage
\clearpage
\setcounter{page}{1}

\twocolumn



\begin{abstract}


In geophysics, hydrocarbon prospect risking involves assessing the risks associated with hydrocarbon exploration by integrating data from various sources. Machine learning-based classifiers trained on tabular data have been recently used to make faster decisions on these prospects. The lack of transparency in the decision-making processes of such models has led to the emergence of explainable AI (XAI). LIME and SHAP are two such examples of these XAI methods which try to generate explanations of a particular decision by ranking the input features in terms of importance.   However, explanations of the same scenario generated by these two different explanation strategies have shown to disagree or be different, particularly for complex data. This is because the definitions of ``importance'' and ``relevance'' differ for different explanation strategies. Thus, grounding these ranked features using theoretically backed causal ideas of necessity and sufficiency can prove to be a more reliable and robust way to improve the trustworthiness of the concerned explanation strategies.We propose a unified framework to generate counterfactuals as well as quantify necessity and sufficiency and use these to perform a robustness evaluation of the explanations provided by LIME and SHAP on high dimensional structured prospect risking data. This robustness test gives us deeper insights into the models capabilities to handle erronous data and which XAI module works best in pair with which model for our dataset for hydorcarbon indication. 

\end{abstract}

\section{Introduction}
\label{introduction}

In high-stakes decision-making tasks within geophysics and seismology, such as interpreting well logs, and assessing subsurface structures, machine learning (ML) classifiers trained on tabular data play a crucial role. Well logs, which provide detailed records of the geological formations encountered during drilling, and seismic data, used to detect hydrocarbon deposits, are often represented in tabular form. This structured format allows for organized and systematic data processing, which is essential for accurate analysis and decision-making. Once such crucial and highly sensitive decision making task involve the analysis of direct hydrocarbon indicators.

Direct hydrocarbon indicators (DHIs) refer to a set of anomalous seismic amplitude patterns that are indicative of the presence of hydrocarbons in subsurface formations. They are caused due to changes in properties of the pore fluid leading to various signatures expressed on the seismic data. Examples of DHIs include bright spots \citep{hammond1974bright}, flat spots \citep{selnes2013flat}, gas chimneys \citep{el2013gas} etc. While DHIs help to reduce uncertainty associated with drilling, they are not infallible and are at times known to produce false positives. Several works in the literature have attempted to analyze the relationship between various DHIs and their ability to successfully predict the presence of hydrocarbons for exploration wells in various amplitude-versus-offset (AVO) classes \citep{roden2012relating, roden2014role}. 

 Since subjectivity among interpreters when grading interpreted DHIs is a major factor in producing false positive decisions, the Rose and Associates DHI consortium \citep{roden2012relating} founded in 2001 attempted to develop a systematic approach to evaluate DHIs. A questionnaire was created whereby each DHI was ranked on a scale from 1 to 5 depending on its exact manifestation in the seismic data, with 5 indicating an almost certain chance of the given DHI leading to a hydrocarbon accumulation and 1 indicating a very low probability in contrast. A database of over 200 drilled prospects was collected with interpreter-evaluated scores of various DHI characteristics for each prospect.
 
 Various works in the literature have attempted to rank DHI attributes in order of importance according to their relationship to successfully drilled prospects \citep{roden2014role, nixon2018ranking}. However, much of this work relies on naively correlating the  frequencies of individual DHIs to the success rate of drilled prospects. This ignores higher order inter-relationships between DHIs that may govern how different subsets of data relate to the probability of success. Additionally, this only reveals general trends that hold on a dataset level as opposed to for particular examples.  
 
 In this regard, the work by \cite{mustafa2022explainable} made an important contribution by leveraging the emerging paradigm of machine learning explainability also referred to as machine learning interpretability or explainable AI (XAI). The field of XAI provides human interpretable analysis in the form of transparent model explanations \citep{alregib2022explanatory}, intuitive visualizations \citep{lee2023probing}, or understandable feature importance scores \citep{mustafa2023explainable} to generate explanations for decisions made by black-box machine learning models. In particular, they showed how these neural networks could be used to learn underlying relationships between DHI attributes to predict prospect outcome and afterwards queried using posthoc machine learning interpretability techniques to explain their decisions for specific prospects. Local interpretability methods are a type of XAI module which have proven to be popular in generating these type of explanations for decision making tasks due to their instance-specific insights into why a particular prediction was made.

Two common local explanation strategies are LIME and SHAP. The basic goal of LIME \citep{ribeiro2016should} is to simulate the behaviour of a complex model, such as a deep neural network, using a model that is simpler and easier to understand, like a linear model. By locally changing the input data and evaluating the ensuing changes in the model's output, LIME seeks to explain specific predictions by ranking features according to importance in producing the outcome. SHAP, on the other hand, \citep{lundberg2017unified} uses feature significance scores, calculated from game theory \citep{vstrumbelj2014explaining}, to categorise each input feature and then uses those scores to explain how machine learning models predict outcomes. These type of methods which give us a feature ranking or score based explanation are also referred to as attribute-based XAI modules.

In spite of all the potential of these XAI strategies, there are discrepancies across well-known XAI methodologies, raising concerns about their reliability \citep{murdoch2019definitions, krishna2022disagreement, kommiya2021towards, ramon2020comparison}. In terms of what `important features' actually means, these attribution based XAI measurements can also be unclear and disagree among themselves. An explanation of the same scenario or prediction generated by LIME and SHAP can give us completely different feature rankings of the same input attributes. Thus, even though different XAI approaches produce different types of interpretation, a convincing explanation should always satisfy the desirable properties of causality: necessity and sufficiency. These two properties address the two questions: \textit{is a feature value necessary for generating the output of the model?}, and \textit{is the feature value sufficient for generating the output of the model?}, respectively. Inclusion of necessity and sufficiency metrics in explanations thus increases trust in explainability approaches \citep{watson2021local, balkir2022necessity, chowdhury2023explaining} and hence, the underlying model results. Essential components are identified by necessity, while sufficiency ensures comprehensive justification. Current methods that formulate these two metrics either require a causal model (which is difficult to obtain) or cannot be implemented on sparse, high dimensional tabular datasets \citep{chowdhury2023counterfactual} which are common in domains like geophysics \citep{mustafa2022explainable} and medicine \citep{prabhushankar2022olives}. 

\citep{kommiya2021towards}, introduced a framework which unify these attribute based methods with another popular XAI strategy called counterfactual based methods to calculate necessity and sufficiency of an input attribute. Counterfactual-based explanations entail creating alternative hypotheses to account for a model's predictions \citep{byrne2007rational, mandel2007psychology, epstude2008functional, prabhushankar2020contrastive, prabhushankar2022introspective}. In order to investigate ``what-if" possibilities, these explanations alter the input attributes and track the model's output changes \citep{van2021conditional}. There are two basic ways: generative models which produce new instances that differ in particular qualities \citep{nemirovsky2021providing, van2021conditional}, and perturbation-based methods \cite{wachter2017counterfactual, mothilal2020explaining, pawelczyk2020learning} that alter input features to produce counterfactuals. Diverse Counterfactual Explanations (DiCE) \citep{mothilal2020explaining}, a common methodology for creating counterfactual explanations, uses optimization techniques to produce a variety of counterfactual examples that are similar to the original instance but have different outcomes.

\begin{figure*}
  \centering
\includegraphics[width=\textwidth,height=0.50\textheight]{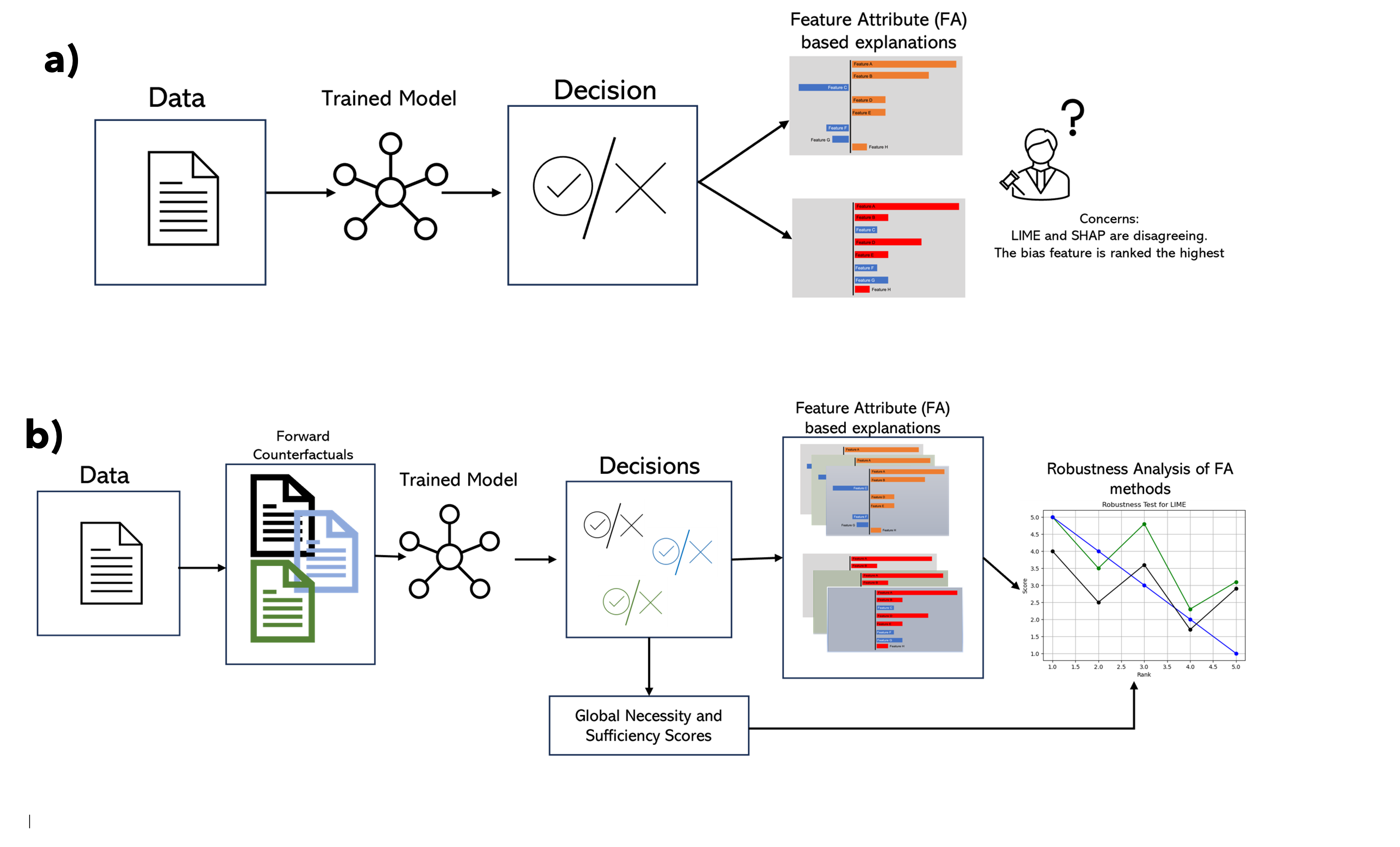} 
  \caption{(a) shows the traditional feature attribute based XAI framework to evaluate local feature importance for a particular decision made on a datapoint by a trained model. (b) is our unified framework to generate forward counterfactuals and global necessity and sufficiency score to do a robustness evaluation (described in Fig x) of given feature attribution (FA) methods (here: LIME and SHAP)}  \label{fig:workflow}
\end{figure*}

In our work we device a unified framework (Fig. \ref{fig:workflow}), based on the methodology provided in \citep{chowdhury2023explaining}, to generate forward counterfactuals and calculate global necessary and sufficiency scores to provide a detailed analysis of the local explanation methods, LIME and SHAP. Our hypothesis is that an important feature should be proportionately necessary and sufficient to generate an outcome. However in our experiments on the DHI data, we find that in particular cases, a feature ranked by LIME or SHAP as more important than others does not rank in the same order in necessity or sufficiency. We have provided further analysis into why it is the case sometimes and described how to chose an explanation when trying to find the most trustworthy decision making models.

In summary, the contribution of this work is three fold:
\begin{enumerate}
    \item A global importance score for each feature set in the DHI dataset based on their necessity and sufficiency to allow comparison with domain knowledge in order to solidify our trust in the method itself.
    \item A unified framework to combine forward counterfactual generation and the global necessity and sufficiency scores to conduct a robustness test for feature attribute based XAI modules on different models trained on a given dataset.
    \item A detailed analysis of the necessity and sufficiency of the important features ranked by two popular local (feature attribute based) explanation methods: LIME and SHAP. We raise and answer the two questions: \textit{Are the important features always necessary?} and \textit{Are the important features sufficient enough?}
\end{enumerate}

\section{Necessity and Sufficiency}
Necessity and sufficiency are concepts that have been extensively explored in philosophy, encompassing various logical, probabilistic, and causal interpretations. \citep{swartz1997concepts} defines necessity and sufficiency as follows:

Necessary condition: A condition $A$ is said to be necessary for a condition $B$, iff the falsity of $A$ guarantees the falsity of $B$.
Example:``Air is necessary for human life."

Sufficient condition: A condition $A$ is said to be sufficient for a condition $B$, iff the truth  of $A$ guarantees the truth of $B$. 
Example:``Being a mammal is sufficient to have a spine."



These two concepts from causal analysis represent the characteristics that one would naturally anticipate the real cause of an event to display \citep{pearl2009causality}. In their work, \citep{kommiya2021towards} have provided the concept of partial explanations that relaxes the necessity and sufficiency conditions to consider the fraction of contexts over which these conditions are valid. Partial explanations are characterized by two metrics.
The conditional probability metrics ($\alpha$) and ($\beta$) respectively captures the extent to which a subset of feature values is either necessary or  sufficiency to cause the model’s (original) output.



\begin{figure*}[t!] 
  \centering
  \includegraphics[width=\textwidth,height=0.55\textheight]{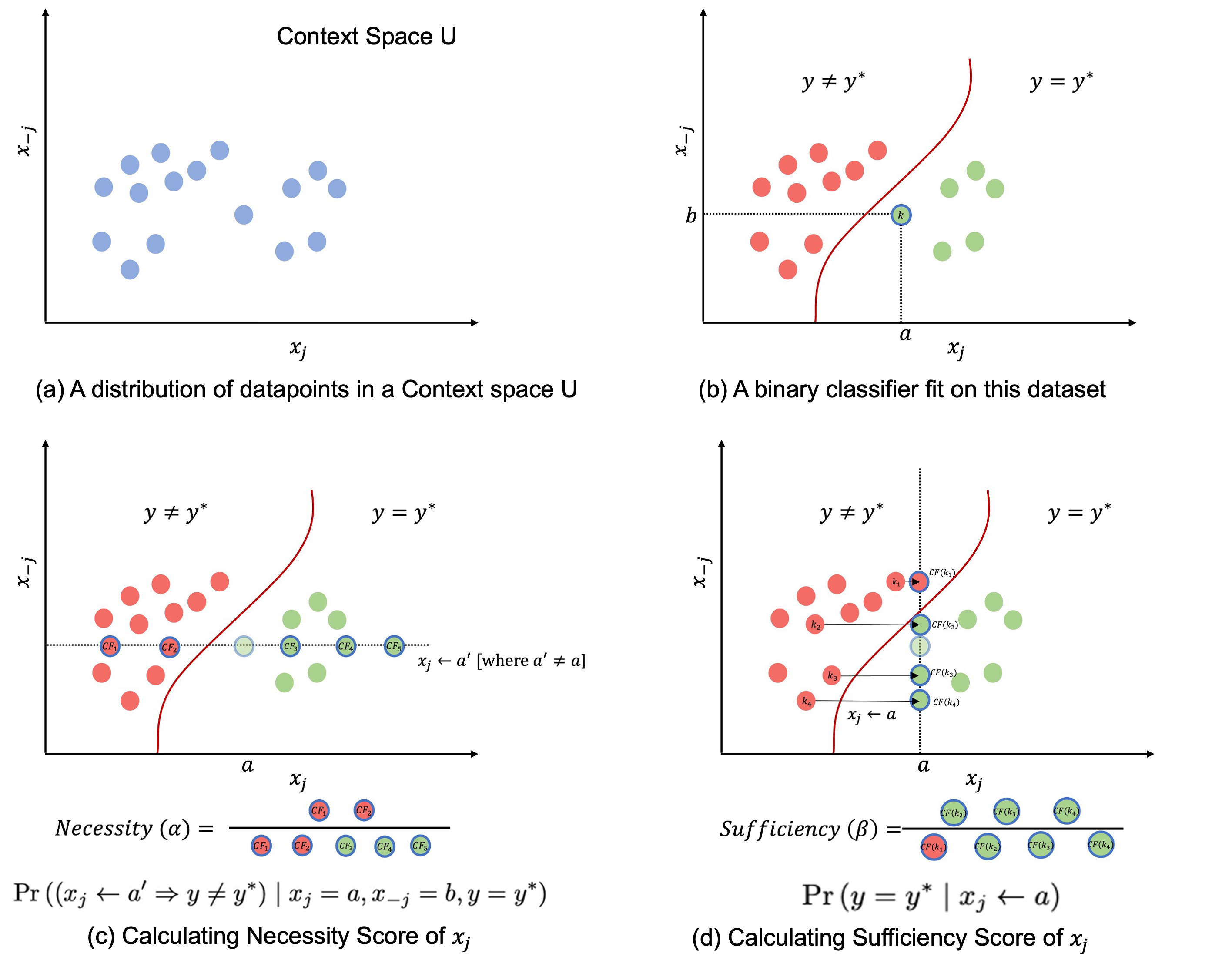} 

  \caption{This Figure explains how necessity and sufficiency calculation for a feature, in a classifier fitted distribution space, works. (a) Shows the distribution is the Context space $U$. This distribution is then fitted to a binary classifier model in (b) while separates the space based on the class prediction of $y=y^* \, \& \, y\neq y^*$. (c) and (d) captures the diagrammatic representation of how intervention is done on the feature $x_j$ to $a'$ and $a$ respectively, to calculate the conditional probability $\alpha$ and $\beta$ based on the change in prediction results.}  \label{fig:nec-surf}

\end{figure*}

Suppose there is a distribution of datapoints in a context space $U$ (Fig.\ref{fig:nec-surf}(a)), where $x_j$ is a subset of feature values relative to the set of context. $x_{-j}$ represents the values of all variables except $x_j$. A binary classifier is fitted on the context space and a reference datapoint $k$ is chosen, whose model output is $y = y^*$ (Fig.\ref{fig:nec-surf}(b)). The initial conditions are: For this datapoint, the desired feature $x_j$ is equal to $a$. While the subset of all other features ($x_{-j}$) from the context $U$ is valued at $b$. Now $x_j$ is intervened and set to any $a'$ [where $a' \neq a$] in (Fig.\ref{fig:nec-surf}(c)). The chances of the model output changing to something other than it's initial state of $y^*$ when the intervention is carried out can thus be represented by:

\begin{equation}\label{eq:necessity}
\alpha=\operatorname{Pr}\left(\left(x_j \leftarrow a^{\prime} \Rightarrow y \neq y^*\right) \mid x_j=a, x_{-j}=b, y=y^*\right)
\end{equation}

Now to calculate $\beta$ we intervene the value of $x_j$ in different datapoints in the context space $U$ in Fig.\ref{fig:nec-surf}(a) to our target value $\alpha$ (from reference point $k$ in Fig.\ref{fig:nec-surf}(b). Thus, the conditional probability of sufficiency is the chances of the intervention, to the target value, leading to the target outcome $y=y^*$.  

\begin{equation}\label{eq:sufficiency}
\beta=\operatorname{Pr}\left(y=y^* \mid x_j \leftarrow a\right)
\end{equation}

Both probabilities $(\alpha, \beta)$, combined called the goodness of an explanation, are over the set of contexts $U$.  When both $\alpha=1$ and $\beta=1$, $\alpha=1$ captures that $x_j=a$ is a necessary cause of $y=y^*$ and $\beta=1$ captures that $x_j=a$ is a sufficient cause of $y=y^*$. In other words, if the subset of feature values $x_j=a$ is an actual cause of the outcome $y=y^*$ with high probability, it is considered to be a good explanation for a model's output.  

To quantify these goodness metrics \citep{kommiya2021towards} suggest a scoring system comprising a necessity score and sufficiency score. Considering $y^*=f\left(x_j=a, x_{-j}=b\right)$ is the output of a classifier $f$ for input $x$. they calculate the necessity score of a feature value $x_j=a$ for the model output $y^*$, by generating counterfactual explanations, but restricting it such that only $x_j$ can be changed. The fraction of times changing $x_j$ leads to the production of a valid DICE or Wachter \citep{wachter2017counterfactual} counterfactual example indicates that the extent to which $x_j=a$ is necessary for the current model output $y^*$.

They adopt the reverse approach for their sufficiency condition. Fixing $x_j$ to its original value, they generated DICE and Wachter Counterfactuals by letting all other features vary their values. The difference between the fraction of unique CFs generated using all features and the fraction of unique CFs generated while keeping $x_j$ constant is their sufficiency score.
However high-dimensional data poses a challenge known as the curse of dimensionality \citep{doshi2017towards, slack2020fooling}. As the number of dimensions increases, the distribution space grows exponentially, resulting in sparse data. This sparsity can make it difficult to generate meaningful CFs using DICE, as there might not be enough instances that match the desired conditions. As the number of dimensions in the data increases, the computational complexity of generating DICE counterfactuals grows significantly as well. Each additional dimension adds to the combinatorial explosion of possibilities, making it computationally expensive and time-consuming to explore all potential counterfactuals.

Furthermore finding erroneous correlations is more likely when there are several dimensions. It can be challenging to pinpoint the precise qualities that are causally responsible for the outcome in high-dimensional datasets because several aspects may be associated with one another. The identification of these causal effects might be made more difficult by the inclusion of confounding variables or hidden factors. Hence it is difficult to come up with a causal structure or model for these datasets, especially without prior knowledge of the domain we are dealing with. Hence the applications of \citep{galhotra2021explaining} in these conditions become very limited as well.
\section{Methodology}
\label{method}

For a counterfactual to be valid and unique, it must fulfil the three conditions:
\begin{enumerate}
    \item causality : it should capture a causal relationship between input features and the model's output
    \item plausibility: it should propose alternative instances that are plausible or realistic in the context of the problem domain and
    \item proximity: it should be close or similar to the original instance in terms of relevant features \citep{wachter2017counterfactual}. As one starts dealing with complex high dimensional data, maintaining all of these conditions become increasingly challenging and thus the option to create a set of valid counterfactual becomes more constrained and limited.
\end{enumerate}

Our method is a model agnostic (flexible when it comes to model selection), post-hoc (calculations are done after the model has been trained) scoring algorithm, which utilizes the predictions generated by the ML model itself, without further approximation. Contrary from DICE counterfactuals, which aims to generate unique counterfactuals, we operationalize Eq.~(\ref{eq:necessity}) and Eq.~(\ref{eq:sufficiency}), and directly calculate both necessity and sufficiency by intervening on a chosen feature ($x_j$) and causing perturbations on the observed datapoints, thus generating these pseudo-counterfactuals which don't necessarily have to be valid and unique or computationally intensive. Also these two equations ((\ref{eq:necessity}) and ~(\ref{eq:sufficiency})) presume that individual feature are independent of each other, allowing any feature to be altered without affecting other features. This allows our algorithm to be functional without a causal model.

\subsection{Counterfactual generation method}
For our algorithm we propose a forward counterfactual generation method, where we directly make perturbations on the selected datapoint, without considering a target prediction. Suppose we have a trained binary classification model, $\mathbf{F}$ (fit on a set of context of tabular data, $\mathbf{U}$). We select an instance, $\mathbf{k}$ with feature vector \(\mathbf{x} = [x_1, x_2, x_3, \ldots, x_n]\) from our context $\mathbf{U}$. Our intention is to generate counterfactual example(s) of this instance $\mathbf{k}$, \{$CF_1(k)$, $CF_2(k)$, …, $CF_n(k)$\}, by introducing fine-grained variations on a single feature value ($x_j$) from the vector space $\mathbf{x}$, keeping all other feature sets constant. A unique counterfactual $CF_i(k)$ can thus be represented in the feature vector space as:
\begin{equation}\label{eq:pseudo-count}
CF_i(k) = [x_1, x_2, \ldots, x_{j-1}, x_j^{(i)}, x_{j+1}, \ldots, x_n]
\end{equation}

Here $x_j^{(i)}$ represents the perturbed value of feature $x_j$ for counterfactual $CF_i(k)$, and all other features are kept the same as their values in $\mathbf{k}$.These newly generated set of counterfactuals are then provided for inference to the trained model $\mathbf{F}$ to obtain it's response

\subsection{Necessity score}
Eq. (\ref{eq:necessity}) can be deduced simply as the following argument: if we can change the model’s output (from $y=y^*$ to $y \neq y^*$) by changing $\mathbf{x_j}$, it means that the $\mathbf{x_j}$ features’ values are necessary to generate the model’s original output ($y=y^*$). Hence we can conclude that for a chosen instance $\mathbf{k}$, the fraction of times a change in $\mathbf{x_j}$, while keeping all other features ($\mathbf{x_j}$) constant, leads to a change in $\mathbf{F}$'s output for the new set of $n$ number of counterfactuals \{$CF_1(k)$, $CF_2(k)$, …, $CF_n(k)$\} is our desired necessity score for feature $\mathbf{x_j}$ (Fig. \ref{fig:nec-surf} c). This scoring is averaged over multiple ($\mathbf{N}$) separate instances in the context space $\mathbf{U}$.
\begin{equation}\label{eq:necessity-ours}
\text { Necessity }=\frac{\sum^{N}{\sum_{i}^{n} \mathbf{1}}\left(CF_i(N) \mid x_j \neq a, y \neq y^*\right)}{\mathrm{n} * N}
\end{equation}

where the model $\mathbf{F}$'s response for the initial instance $\mathbf{k}$ was $y=y^*$

\subsection{Sufficiency score}

Similarly as above, Eq (\ref{eq:sufficiency}) can be explained in the following way:
if intervening $\mathbf{x_j}$ to a certain value of $a$ leads us to a targeted output by the model (from $y \neq y^*$  to $y=y^*$ ), then we can conclude that $\mathbf{x_j}$ is sufficient to generate the model's desired output ($y=y^*$).To begin formulating our equation for sufficiency score, we first need to select our intervention value for $\mathbf{x_j}$ and desired outcome of our model $\mathbf{F}$. To do this, we choose a reference datapoint $\mathbf{r}$ from our context space $\mathbf{U}$ (Fig. \ref{fig:nec-surf} b). The value of feature $\mathbf{x_j}$ of $\mathbf{r}$ (suppose: $x_j=a$) becomes our intervention value for the corresponding feature and the model response to $\mathbf{r}$ is our desired/targeted outcome. (suppose: $y= \mathbf{F}(r) = y^*$).

Now, to draw our sample space to do our intervention for Eq. (\ref{eq:sufficiency}) we select a set of instances $\mathbf{K}$, which produces the undesired model response ($y \neq y^*$) from our test observational set. This is done to ensure that during intervention the newly generated counterfactual does not become too much of an outlier for our context $U$. For each $\mathbf{k} \in\mathbf{K}$, we intervene the value of $\mathbf{x_j}$ to $a$ (Fig. \ref{fig:nec-surf} d).

Thus, the sufficiency score of feature $\mathbf{x_j}$ for reference point $\mathbf{r}$ can be computed as the fraction of counterfactuals ($CF(k)$) generated from the set $\mathbf{K}$, producing the targeted outcome ($y=y^*$) from an undesired one ($y \neq y^*$) when $\mathbf{x_j}$ is intervened to value $a$ (${x}_j \leftarrow a$). This scoring is averaged over multiple ($\mathbf{R}$) references in the context space $\mathbf{U}$.

\begin{equation}\label{eq:sufficiency-ours}
\text { Sufficiency }=\frac{\sum^{R}{\sum^{K} \mathbf{1}}\left(CF(k) \mid {x}_j \leftarrow a, y=y^*\right)}{\mathrm{K} * R}
\end{equation}

where $k \in K$ and the model response $\mathbf{F(\forall k)}$ was the undesired one ($y \neq y^*$).

Using these two scores we come up with our own global importance scoring metric based on overall necessity and sufficiency of a particular feature set in an effort to establish the reliability of our algorithm by anchoring it to domain knowledge.

\subsection{Global Feature Importance}

To calculate this we follow the method provided in \citep{mustafa2022explainable} to summarize global model behavior through aggregation of local responses.

\begin{equation}
{\Gamma}_{\text{global}}^i = \frac{1}{|\mathcal{U}|} \sum_{k \in \mathcal{U}} \left|\gamma^i{(k)}\right|
\end{equation}

Where $\Gamma$ and $\gamma$ are the global and local (necessity/sufficiency) score respectively, for a feature $i$. $k$ is an element in the context space $\mathcal{U}$. 
$\Gamma^i_\text{global}$ is calculates by averaging the local score $\gamma^i$ over all elements of $\mathcal{U}$.

In the next set of experiments we evaluate the explanations provided by two attribution based feature importance methods, LIME and SHAP, in different setups.

\section{Experiments}

We conduct two separate sets of experiments using our scoring metrics. First, is to generate a global necessity and sufficiency values for each of the 6 high level features present in the dataset. And next is to conduct a necessity and sufficiency analysis of the features obtained as results of LIME and SHAP explanation.

For our experiments, we have divided the 35 attributes in the dataset into two corresponding groups: the Top 7 high level features and the rest of the features.

To begin, we need to select our classifier models to calculate our scores upon. We have used a Logistic Regression model, a Gaussian Naive Bayes, a Random Forest Classifier and a Voting Classifier (which considers the weight of the previous three models in a $1: 5: 1$ proportion). All these models are selected from the popular sklearn library available for basic machine learning tasks in python.

We split our dataset into a $70:30$ ratio to provide our training set and testing set (on which the interventions will be carried out) respectively. Table (\ref{tab:accuracies1}) reports the accuracy results for both the training and testing of these classifiers on the above mentioned data-split.

\begin{table}[ht]
\centering
\resizebox{\columnwidth}{!}{%
\begin{tabular}{|c|cc|cc|}
\hline
 &
  \multicolumn{2}{c|}{With all features} &
  \multicolumn{2}{c|}{With seven high level features} \\ \cline{2-5} 
\multirow{-2}{*}{Model} &
  \multicolumn{1}{c|}{\textit{Training Accuracy}} &
  \textit{Validation Accuracy} &
  \multicolumn{1}{c|}{\textit{Training Accuracy}} &
  \textit{Validation Accuracy} \\ \hline
\textbf{Logistic Regression} &
  \multicolumn{1}{c|}{{\color[HTML]{333333} 0.7949}} &
  {\color[HTML]{333333} 0.7327} &
  \multicolumn{1}{c|}{{\color[HTML]{333333} 0.7906}} &
  {\color[HTML]{333333} 0.7129} \\ \hline
\textbf{Gaussian NB} &
  \multicolumn{1}{c|}{{\color[HTML]{333333} 0.7778}} &
  {\color[HTML]{333333} 0.7329} &
  \multicolumn{1}{c|}{{\color[HTML]{333333} 0.7991}} &
  {\color[HTML]{333333} 0.7426} \\ \hline
\textbf{Random Forests} &
  \multicolumn{1}{c|}{{\color[HTML]{333333} 1.00}} &
  {\color[HTML]{333333} 0.7525} &
  \multicolumn{1}{c|}{{\color[HTML]{333333} 1.00}} &
  {\color[HTML]{333333} 0.7228} \\ \hline
\textbf{Voting Classifier (1:5:1)} &
  \multicolumn{1}{c|}{{\color[HTML]{333333} 0.7821}} &
  {\color[HTML]{333333} 0.7330} &
  \multicolumn{1}{c|}{{\color[HTML]{333333} 0.8034}} &
  {\color[HTML]{333333} 0.7426} \\ \hline
\end{tabular}%
}
\caption{The training and the test accuracies of the binary classifiers used for the experiments}
\label{tab:accuracies1}
\end{table}

\subsection{Robustness evaluation of LIME and SHAPLEY}

In this section we use our scores to examine the necessity and sufficiency of the top features as declared by the two feature attribution methods, namely LIME and SHAP. This means that we use Eq. (\ref{eq:necessity-ours} \& \ref{eq:sufficiency-ours}) to calculate the necessity and sufficiency scores of the $k$-th features in a LIME or SHAP explanation of a particular datapoint. Thus the $k$-th feature becomes out intervention feature ($x_j$) for that concerned datapoint. 

Provided the top ``important" features identified based on these attribution methods, we compare the necessity and sufficiency scores of the the top $k$-th most important features ($k \in\{1,2,3, \text{rest}\}$) with each other. Specifically we measure the average scores of these top $k$ features, ranked by LIME and SHAP, over a span of multiple separate test datapoints. ($N$ \& $R$ in Eq. (\ref{eq:necessity-ours} \& \ref{eq:sufficiency-ours})).

We need to remember that all these scores are defined with respect to the feature value ($a$) and original output prediction ($y$) of each of these individual datapoints.

Practically no two same features are in the same importance ranking of two different datapoints, even by the same explanation method. Hence it raises the questions: Are the important features always necessary? and, Are the important features sufficient?. 

We hypothesize that: ``In an ideal setting, for an explanation to be robust, the features ranked by them in their own scale of importance show also be proportionately necessary and sufficient at the same time, i.e. the most important feature should be both the most necessary one and the most sufficient one with necessity and sufficiency scores decreasing as their ranks decrease."
\section{Conclusion}

The global necessity and sufficiency scores we provide help interpreters and domain experts working with this DHI or similar data to validate our metrics using their knowledge. Establishing association between an expert's way of making a certain decision and the technique by which a model selects and chooses a necessary or sufficient features to make a prediction will lead to a better understanding of the inner workings of the classifiers and ultimately will translate to higher levels of trust between the user and these ML models. The necessity and sufficiency evaluation of LIME and SHAP explanations of the high dimensional structured DHI data gives insights into how these different strategies behave when generating explanations of different forms of complex data and when used on different types of ML models. It helps us to select the most robust explanation method in a particular scenario and for a particular model thus leading to a more trustworthy process of decision making and evaluation. 

In conclusion our study emphasises the value of employing several explanation techniques with a combination of different ml models, and also provides a proper evaluation method to determine which explanation technique and which model are most suitable for a particular task. The temptation to view a particular explanation as a one-size-fits-all solution must be resisted by our scientific community. Overall, we think that in order to make a trustworthy and informed decision about the behavior of an ML model the employment of several explanation methods backed by theoretical concepts and domain knowledge will only prove to be useful.




\appendix

\section{Appendix}\label{sec:appen1}

We have listed here a table of of the top-$k$ occurrences in LIME and SHAP explanations of each of the top 7 features \{ Anomaly Class, Initial Pg, DHI index, calibrated Pg, and data quality scores for composite, seismic, and rock and fluid data \} in the DHI dataset. 

\subsubsection{Top $k$ occurrence for LIME Explanations:}
\begin{table}[H]
\centering
\resizebox{\columnwidth}{!}{%
\begin{tabular}{|ccccccccc|}
\hline
\multicolumn{9}{|c|}{{\color[HTML]{333333} \textbf{Top k-th Occurrences of high level features in LIME explanations (Model = Logistic Regression}}} \\ \hline
\multicolumn{1}{|c|}{{\color[HTML]{333333} }} &
  \multicolumn{4}{c|}{{\color[HTML]{333333} \textit{\textbf{For All Features}}}} &
  \multicolumn{4}{c|}{{\color[HTML]{333333} \textit{\textbf{For Top 7 Features}}}} \\ \cline{2-9} 
\multicolumn{1}{|c|}{\multirow{-2}{*}{{\color[HTML]{333333} \textbf{\begin{tabular}[c]{@{}c@{}}Feature\\ Name\end{tabular}}}}} &
  \multicolumn{1}{c|}{{\color[HTML]{333333} \textit{\textbf{\begin{tabular}[c]{@{}c@{}}Top 1\\ occurrence\end{tabular}}}}} &
  \multicolumn{1}{c|}{{\color[HTML]{333333} \textit{\textbf{\begin{tabular}[c]{@{}c@{}}Top 2\\ occurrence\end{tabular}}}}} &
  \multicolumn{1}{c|}{{\color[HTML]{333333} \textit{\textbf{\begin{tabular}[c]{@{}c@{}}Top 3\\ occurrence\end{tabular}}}}} &
  \multicolumn{1}{c|}{{\color[HTML]{333333} \textit{\textbf{\begin{tabular}[c]{@{}c@{}}Top 5\\ occurrence\end{tabular}}}}} &
  \multicolumn{1}{c|}{{\color[HTML]{333333} \textit{\textbf{\begin{tabular}[c]{@{}c@{}}Top 1\\ occurrence\end{tabular}}}}} &
  \multicolumn{1}{c|}{{\color[HTML]{333333} \textit{\textbf{\begin{tabular}[c]{@{}c@{}}Top 2\\ occurrence\end{tabular}}}}} &
  \multicolumn{1}{c|}{{\color[HTML]{333333} \textit{\textbf{\begin{tabular}[c]{@{}c@{}}Top 3\\ occurrence\end{tabular}}}}} &
  {\color[HTML]{333333} \textit{\textbf{\begin{tabular}[c]{@{}c@{}}Top 5\\ occurrence\end{tabular}}}} \\ \hline
\multicolumn{1}{|c|}{{\color[HTML]{333333} \textbf{Feature 1}}} &
  \multicolumn{1}{c|}{{\color[HTML]{333333} 0}} &
  \multicolumn{1}{c|}{{\color[HTML]{333333} 0}} &
  \multicolumn{1}{c|}{{\color[HTML]{333333} 0}} &
  \multicolumn{1}{c|}{{\color[HTML]{333333} 1}} &
  \multicolumn{1}{c|}{{\color[HTML]{333333} 1}} &
  \multicolumn{1}{c|}{{\color[HTML]{333333} 96}} &
  \multicolumn{1}{c|}{{\color[HTML]{333333} 101}} &
  {\color[HTML]{333333} 101} \\ \hline
\multicolumn{1}{|c|}{{\color[HTML]{333333} \textbf{Feature 2}}} &
  \multicolumn{1}{c|}{{\color[HTML]{333333} 0}} &
  \multicolumn{1}{c|}{{\color[HTML]{333333} 0}} &
  \multicolumn{1}{c|}{{\color[HTML]{333333} 0}} &
  \multicolumn{1}{c|}{{\color[HTML]{333333} 0}} &
  \multicolumn{1}{c|}{{\color[HTML]{333333} 0}} &
  \multicolumn{1}{c|}{{\color[HTML]{333333} \textbf{5}}} &
  \multicolumn{1}{c|}{{\color[HTML]{CB0000} \textbf{61}}} &
  {\color[HTML]{333333} 91} \\ \hline
\multicolumn{1}{|c|}{{\color[HTML]{333333} \textbf{\begin{tabular}[c]{@{}c@{}}Feature 3\end{tabular}}}} &
  \multicolumn{1}{c|}{{\color[HTML]{333333} 0}} &
  \multicolumn{1}{c|}{{\color[HTML]{333333} 0}} &
  \multicolumn{1}{c|}{{\color[HTML]{333333} 0}} &
  \multicolumn{1}{c|}{{\color[HTML]{333333} 0}} &
  \multicolumn{1}{c|}{{\color[HTML]{333333} 0}} &
  \multicolumn{1}{c|}{{\color[HTML]{333333} \textbf{0}}} &
  \multicolumn{1}{c|}{{\color[HTML]{CB0000} \textbf{27}}} &
  {\color[HTML]{333333} 77} \\ \hline
\multicolumn{1}{|c|}{{\color[HTML]{333333} \textbf{Feature 4}}} &
  \multicolumn{1}{c|}{{\color[HTML]{333333} 49}} &
  \multicolumn{1}{c|}{{\color[HTML]{333333} 49}} &
  \multicolumn{1}{c|}{{\color[HTML]{333333} 50}} &
  \multicolumn{1}{c|}{{\color[HTML]{333333} 57}} &
  \multicolumn{1}{c|}{{\color[HTML]{333333} 100}} &
  \multicolumn{1}{c|}{{\color[HTML]{333333} 101}} &
  \multicolumn{1}{c|}{{\color[HTML]{333333} 101}} &
  {\color[HTML]{333333} 101} \\ \hline
\multicolumn{1}{|c|}{{\color[HTML]{333333} \textbf{\begin{tabular}[c]{@{}c@{}}Feature 5\end{tabular}}}} &
  \multicolumn{1}{c|}{{\color[HTML]{333333} 0}} &
  \multicolumn{1}{c|}{{\color[HTML]{333333} 0}} &
  \multicolumn{1}{c|}{{\color[HTML]{333333} 0}} &
  \multicolumn{1}{c|}{{\color[HTML]{333333} 0}} &
  \multicolumn{1}{c|}{{\color[HTML]{333333} 0}} &
  \multicolumn{1}{c|}{{\color[HTML]{333333} 0}} &
  \multicolumn{1}{c|}{{\color[HTML]{333333} 0}} &
  {\color[HTML]{333333} 11} \\ \hline
\multicolumn{1}{|c|}{{\color[HTML]{333333} \textbf{\begin{tabular}[c]{@{}c@{}}Feature 6\end{tabular}}}} &
  \multicolumn{1}{c|}{{\color[HTML]{333333} 0}} &
  \multicolumn{1}{c|}{{\color[HTML]{333333} 0}} &
  \multicolumn{1}{c|}{{\color[HTML]{333333} 0}} &
  \multicolumn{1}{c|}{{\color[HTML]{333333} 0}} &
  \multicolumn{1}{c|}{{\color[HTML]{333333} 0}} &
  \multicolumn{1}{c|}{{\color[HTML]{333333} 0}} &
  \multicolumn{1}{c|}{{\color[HTML]{333333} 8}} &
  {\color[HTML]{333333} 66} \\ \hline
\multicolumn{1}{|c|}{{\color[HTML]{333333} \textbf{\begin{tabular}[c]{@{}c@{}}Feature 7\end{tabular}}}} &
  \multicolumn{1}{c|}{{\color[HTML]{333333} 0}} &
  \multicolumn{1}{c|}{{\color[HTML]{333333} 0}} &
  \multicolumn{1}{c|}{{\color[HTML]{333333} 0}} &
  \multicolumn{1}{c|}{{\color[HTML]{333333} 0}} &
  \multicolumn{1}{c|}{{\color[HTML]{333333} 0}} &
  \multicolumn{1}{c|}{{\color[HTML]{333333} 0}} &
  \multicolumn{1}{c|}{{\color[HTML]{333333} 5}} &
  {\color[HTML]{333333} 58} \\ \hline
\multicolumn{1}{|c|}{{\color[HTML]{333333} \textbf{Rest (Averaged)}}} &
  \multicolumn{1}{c|}{{\color[HTML]{333333} 2.260869565}} &
  \multicolumn{1}{c|}{{\color[HTML]{333333} 6.652173913}} &
  \multicolumn{1}{c|}{{\color[HTML]{333333} 11}} &
  \multicolumn{1}{c|}{{\color[HTML]{333333} 19.43478261}} &
  \multicolumn{1}{c|}{{\color[HTML]{333333} -}} &
  \multicolumn{1}{c|}{{\color[HTML]{333333} -}} &
  \multicolumn{1}{c|}{{\color[HTML]{333333} -}} &
  {\color[HTML]{333333} -} \\ \hline
\end{tabular}%
}
\caption{Top $k$ occurrence for the Top 7 features of the DHI dataset as ranked by LIME during explanation of 101 test cases for model = Logistic Regression.}

\label{tab:lime-lr-rank}
\end{table}

\begin{table}[H]
\centering
\resizebox{\columnwidth}{!}{%
\begin{tabular}{|ccccccccc|}
\hline
\multicolumn{9}{|c|}{{\color[HTML]{333333} \textbf{Top k-th Occurrences of high level features in LIME explanations (Model= Gaussian NB)}}} \\ \hline
\multicolumn{1}{|c|}{{\color[HTML]{333333} }} &
  \multicolumn{4}{c|}{{\color[HTML]{333333} \textit{\textbf{For All Features}}}} &
  \multicolumn{4}{c|}{{\color[HTML]{333333} \textit{\textbf{For Top 7 Features}}}} \\ \cline{2-9} 
\multicolumn{1}{|c|}{\multirow{-2}{*}{{\color[HTML]{333333} \textbf{\begin{tabular}[c]{@{}c@{}}Feature\\ Name\end{tabular}}}}} &
  \multicolumn{1}{c|}{{\color[HTML]{333333} \textit{\textbf{\begin{tabular}[c]{@{}c@{}}Top 1\\ occurrence\end{tabular}}}}} &
  \multicolumn{1}{c|}{{\color[HTML]{333333} \textit{\textbf{\begin{tabular}[c]{@{}c@{}}Top 2\\ occurrence\end{tabular}}}}} &
  \multicolumn{1}{c|}{{\color[HTML]{333333} \textit{\textbf{\begin{tabular}[c]{@{}c@{}}Top 3\\ occurrence\end{tabular}}}}} &
  \multicolumn{1}{c|}{{\color[HTML]{333333} \textit{\textbf{\begin{tabular}[c]{@{}c@{}}Top 5\\ occurrence\end{tabular}}}}} &
  \multicolumn{1}{c|}{{\color[HTML]{333333} \textit{\textbf{\begin{tabular}[c]{@{}c@{}}Top 1\\ occurrence\end{tabular}}}}} &
  \multicolumn{1}{c|}{{\color[HTML]{333333} \textit{\textbf{\begin{tabular}[c]{@{}c@{}}Top 2\\ occurrence\end{tabular}}}}} &
  \multicolumn{1}{c|}{{\color[HTML]{333333} \textit{\textbf{\begin{tabular}[c]{@{}c@{}}Top 3\\ occurrence\end{tabular}}}}} &
  {\color[HTML]{333333} \textit{\textbf{\begin{tabular}[c]{@{}c@{}}Top 5\\ occurrence\end{tabular}}}} \\ \hline
\multicolumn{1}{|c|}{{\color[HTML]{333333} \textbf{Feature 1}}} &
  \multicolumn{1}{c|}{{\color[HTML]{333333} 4}} &
  \multicolumn{1}{c|}{{\color[HTML]{333333} 13}} &
  \multicolumn{1}{c|}{{\color[HTML]{333333} 29}} &
  \multicolumn{1}{c|}{{\color[HTML]{333333} 53}} &
  \multicolumn{1}{c|}{{\color[HTML]{333333} 20}} &
  \multicolumn{1}{c|}{{\color[HTML]{333333} 43}} &
  \multicolumn{1}{c|}{{\color[HTML]{333333} 81}} &
  {\color[HTML]{333333} 101} \\ \hline
\multicolumn{1}{|c|}{{\color[HTML]{333333} \textbf{Feature 2}}} &
  \multicolumn{1}{c|}{{\color[HTML]{333333} 10}} &
  \multicolumn{1}{c|}{{\color[HTML]{333333} 15}} &
  \multicolumn{1}{c|}{{\color[HTML]{333333} \textbf{27}}} &
  \multicolumn{1}{c|}{{\color[HTML]{333333} \textbf{52}}} &
  \multicolumn{1}{c|}{{\color[HTML]{333333} 10}} &
  \multicolumn{1}{c|}{{\color[HTML]{333333} 19}} &
  \multicolumn{1}{c|}{{\color[HTML]{333333} \textbf{47}}} &
  {\color[HTML]{333333} \textbf{76}} \\ \hline
\multicolumn{1}{|c|}{{\color[HTML]{333333} \textbf{\begin{tabular}[c]{@{}c@{}}Feature 3\end{tabular}}}} &
  \multicolumn{1}{c|}{{\color[HTML]{CB0000} \textbf{6}}} &
  \multicolumn{1}{c|}{{\color[HTML]{CB0000} \textbf{44}}} &
  \multicolumn{1}{c|}{{\color[HTML]{333333} \textbf{49}}} &
  \multicolumn{1}{c|}{{\color[HTML]{333333} \textbf{63}}} &
  \multicolumn{1}{c|}{{\color[HTML]{CB0000} \textbf{2}}} &
  \multicolumn{1}{c|}{{\color[HTML]{CB0000} \textbf{49}}} &
  \multicolumn{1}{c|}{{\color[HTML]{333333} \textbf{72}}} &
  {\color[HTML]{333333} \textbf{98}} \\ \hline
\multicolumn{1}{|c|}{{\color[HTML]{333333} \textbf{Feature 4}}} &
  \multicolumn{1}{c|}{{\color[HTML]{333333} 48}} &
  \multicolumn{1}{c|}{{\color[HTML]{333333} 61}} &
  \multicolumn{1}{c|}{{\color[HTML]{333333} 77}} &
  \multicolumn{1}{c|}{{\color[HTML]{333333} 96}} &
  \multicolumn{1}{c|}{{\color[HTML]{333333} 69}} &
  \multicolumn{1}{c|}{{\color[HTML]{333333} 91}} &
  \multicolumn{1}{c|}{{\color[HTML]{333333} 100}} &
  {\color[HTML]{333333} 101} \\ \hline
\multicolumn{1}{|c|}{{\color[HTML]{333333} \textbf{\begin{tabular}[c]{@{}c@{}}Feature 5\end{tabular}}}} &
  \multicolumn{1}{c|}{{\color[HTML]{333333} 0}} &
  \multicolumn{1}{c|}{{\color[HTML]{333333} 0}} &
  \multicolumn{1}{c|}{{\color[HTML]{333333} 0}} &
  \multicolumn{1}{c|}{{\color[HTML]{333333} 0}} &
  \multicolumn{1}{c|}{{\color[HTML]{333333} 0}} &
  \multicolumn{1}{c|}{{\color[HTML]{333333} 0}} &
  \multicolumn{1}{c|}{{\color[HTML]{333333} 0}} &
  {\color[HTML]{333333} 35} \\ \hline
\multicolumn{1}{|c|}{{\color[HTML]{333333} \textbf{\begin{tabular}[c]{@{}c@{}}Feature 6\end{tabular}}}} &
  \multicolumn{1}{c|}{{\color[HTML]{333333} 0}} &
  \multicolumn{1}{c|}{{\color[HTML]{333333} 0}} &
  \multicolumn{1}{c|}{{\color[HTML]{333333} 0}} &
  \multicolumn{1}{c|}{{\color[HTML]{333333} 0}} &
  \multicolumn{1}{c|}{{\color[HTML]{333333} 0}} &
  \multicolumn{1}{c|}{{\color[HTML]{333333} 0}} &
  \multicolumn{1}{c|}{{\color[HTML]{333333} 0}} &
  {\color[HTML]{333333} 8} \\ \hline
\multicolumn{1}{|c|}{{\color[HTML]{333333} \textbf{\begin{tabular}[c]{@{}c@{}}Feature 7\end{tabular}}}} &
  \multicolumn{1}{c|}{{\color[HTML]{333333} 0}} &
  \multicolumn{1}{c|}{{\color[HTML]{333333} 0}} &
  \multicolumn{1}{c|}{{\color[HTML]{333333} 1}} &
  \multicolumn{1}{c|}{{\color[HTML]{333333} 9}} &
  \multicolumn{1}{c|}{{\color[HTML]{333333} 0}} &
  \multicolumn{1}{c|}{{\color[HTML]{333333} 0}} &
  \multicolumn{1}{c|}{{\color[HTML]{333333} 3}} &
  {\color[HTML]{333333} 86} \\ \hline
\multicolumn{1}{|c|}{{\color[HTML]{333333} \textbf{Rest (Averaged)}}} &
  \multicolumn{1}{c|}{{\color[HTML]{333333} 1.43478261}} &
  \multicolumn{1}{c|}{{\color[HTML]{333333} 3}} &
  \multicolumn{1}{c|}{{\color[HTML]{333333} 5.2173913}} &
  \multicolumn{1}{c|}{{\color[HTML]{333333} 10.0869565}} &
  \multicolumn{1}{c|}{{\color[HTML]{333333} -}} &
  \multicolumn{1}{c|}{{\color[HTML]{333333} -}} &
  \multicolumn{1}{c|}{{\color[HTML]{333333} -}} &
  {\color[HTML]{333333} -} \\ \hline
\end{tabular}%
}

\caption{Top $k$ occurrence for the top seven features of the DHI dataset as ranked by LIME during explanation of 101 test cases for model = Gaussian NB}
\label{tab:lime-gnb-rank}

\end{table}

\begin{table}[H]
\centering
\resizebox{\columnwidth}{!}{%
\begin{tabular}{|ccccccccc|}
\hline
\multicolumn{9}{|c|}{{\color[HTML]{333333} \textbf{Top k-th Occurrences of high level features in LIME explanations (Model= Random Forest)}}} \\ \hline
\multicolumn{1}{|c|}{{\color[HTML]{333333} }} &
  \multicolumn{4}{c|}{{\color[HTML]{333333} \textit{\textbf{For All Features}}}} &
  \multicolumn{4}{c|}{{\color[HTML]{333333} \textit{\textbf{For Top 7 Features}}}} \\ \cline{2-9} 
\multicolumn{1}{|c|}{\multirow{-2}{*}{{\color[HTML]{333333} \textbf{\begin{tabular}[c]{@{}c@{}}Feature\\ Name\end{tabular}}}}} &
  \multicolumn{1}{c|}{{\color[HTML]{333333} \textit{\textbf{\begin{tabular}[c]{@{}c@{}}Top 1\\ occurrence\end{tabular}}}}} &
  \multicolumn{1}{c|}{{\color[HTML]{333333} \textit{\textbf{\begin{tabular}[c]{@{}c@{}}Top 2\\ occurrence\end{tabular}}}}} &
  \multicolumn{1}{c|}{{\color[HTML]{333333} \textit{\textbf{\begin{tabular}[c]{@{}c@{}}Top 3\\ occurrence\end{tabular}}}}} &
  \multicolumn{1}{c|}{{\color[HTML]{333333} \textit{\textbf{\begin{tabular}[c]{@{}c@{}}Top 5\\ occurrence\end{tabular}}}}} &
  \multicolumn{1}{c|}{{\color[HTML]{333333} \textit{\textbf{\begin{tabular}[c]{@{}c@{}}Top 1\\ occurrence\end{tabular}}}}} &
  \multicolumn{1}{c|}{{\color[HTML]{333333} \textit{\textbf{\begin{tabular}[c]{@{}c@{}}Top 2\\ occurrence\end{tabular}}}}} &
  \multicolumn{1}{c|}{{\color[HTML]{333333} \textit{\textbf{\begin{tabular}[c]{@{}c@{}}Top 3\\ occurrence\end{tabular}}}}} &
  {\color[HTML]{333333} \textit{\textbf{\begin{tabular}[c]{@{}c@{}}Top 5\\ occurrence\end{tabular}}}} \\ \hline
\multicolumn{1}{|c|}{{\color[HTML]{333333} \textbf{Feature 1}}} &
  \multicolumn{1}{c|}{{\color[HTML]{333333} 1}} &
  \multicolumn{1}{c|}{{\color[HTML]{333333} 1}} &
  \multicolumn{1}{c|}{{\color[HTML]{333333} 7}} &
  \multicolumn{1}{c|}{{\color[HTML]{333333} 23}} &
  \multicolumn{1}{c|}{{\color[HTML]{333333} 0}} &
  \multicolumn{1}{c|}{{\color[HTML]{333333} 2}} &
  \multicolumn{1}{c|}{{\color[HTML]{333333} 16}} &
  {\color[HTML]{333333} 78} \\ \hline
\multicolumn{1}{|c|}{{\color[HTML]{333333} \textbf{Feature 2}}} &
  \multicolumn{1}{c|}{{\color[HTML]{333333} 24}} &
  \multicolumn{1}{c|}{{\color[HTML]{333333} 34}} &
  \multicolumn{1}{c|}{{\color[HTML]{333333} 80}} &
  \multicolumn{1}{c|}{{\color[HTML]{333333} 93}} &
  \multicolumn{1}{c|}{{\color[HTML]{333333} 0}} &
  \multicolumn{1}{c|}{{\color[HTML]{333333} 42}} &
  \multicolumn{1}{c|}{{\color[HTML]{333333} 82}} &
  {\color[HTML]{333333} 101} \\ \hline
\multicolumn{1}{|c|}{{\color[HTML]{333333} \textbf{\begin{tabular}[c]{@{}c@{}}Feature 3\end{tabular}}}} &
  \multicolumn{1}{c|}{{\color[HTML]{333333} 48}} &
  \multicolumn{1}{c|}{{\color[HTML]{333333} 70}} &
  \multicolumn{1}{c|}{{\color[HTML]{333333} 71}} &
  \multicolumn{1}{c|}{{\color[HTML]{333333} 71}} &
  \multicolumn{1}{c|}{{\color[HTML]{333333} 1}} &
  \multicolumn{1}{c|}{{\color[HTML]{333333} 52}} &
  \multicolumn{1}{c|}{{\color[HTML]{333333} 75}} &
  {\color[HTML]{333333} 80} \\ \hline
\multicolumn{1}{|c|}{{\color[HTML]{333333} \textbf{Feature 4}}} &
  \multicolumn{1}{c|}{{\color[HTML]{333333} 25}} &
  \multicolumn{1}{c|}{{\color[HTML]{333333} 76}} &
  \multicolumn{1}{c|}{{\color[HTML]{333333} 94}} &
  \multicolumn{1}{c|}{{\color[HTML]{333333} 101}} &
  \multicolumn{1}{c|}{{\color[HTML]{333333} 100}} &
  \multicolumn{1}{c|}{{\color[HTML]{333333} 101}} &
  \multicolumn{1}{c|}{{\color[HTML]{333333} 101}} &
  {\color[HTML]{333333} 101} \\ \hline
\multicolumn{1}{|c|}{{\color[HTML]{333333} \textbf{\begin{tabular}[c]{@{}c@{}}Feature 5\end{tabular}}}} &
  \multicolumn{1}{c|}{{\color[HTML]{333333} 0}} &
  \multicolumn{1}{c|}{{\color[HTML]{333333} 0}} &
  \multicolumn{1}{c|}{{\color[HTML]{333333} 0}} &
  \multicolumn{1}{c|}{{\color[HTML]{333333} 1}} &
  \multicolumn{1}{c|}{{\color[HTML]{333333} 0}} &
  \multicolumn{1}{c|}{{\color[HTML]{333333} 0}} &
  \multicolumn{1}{c|}{{\color[HTML]{333333} 0}} &
  {\color[HTML]{333333} 21} \\ \hline
\multicolumn{1}{|c|}{{\color[HTML]{333333} \textbf{\begin{tabular}[c]{@{}c@{}}Feature 6\end{tabular}}}} &
  \multicolumn{1}{c|}{{\color[HTML]{333333} 0}} &
  \multicolumn{1}{c|}{{\color[HTML]{333333} 0}} &
  \multicolumn{1}{c|}{{\color[HTML]{333333} 0}} &
  \multicolumn{1}{c|}{{\color[HTML]{333333} 0}} &
  \multicolumn{1}{c|}{{\color[HTML]{333333} 0}} &
  \multicolumn{1}{c|}{{\color[HTML]{333333} 5}} &
  \multicolumn{1}{c|}{{\color[HTML]{333333} 28}} &
  {\color[HTML]{333333} 82} \\ \hline
\multicolumn{1}{|c|}{{\color[HTML]{333333} \textbf{\begin{tabular}[c]{@{}c@{}}Feature 7\end{tabular}}}} &
  \multicolumn{1}{c|}{{\color[HTML]{333333} 0}} &
  \multicolumn{1}{c|}{{\color[HTML]{333333} 0}} &
  \multicolumn{1}{c|}{{\color[HTML]{333333} 0}} &
  \multicolumn{1}{c|}{{\color[HTML]{333333} 1}} &
  \multicolumn{1}{c|}{{\color[HTML]{333333} 0}} &
  \multicolumn{1}{c|}{{\color[HTML]{333333} 0}} &
  \multicolumn{1}{c|}{{\color[HTML]{333333} 1}} &
  {\color[HTML]{333333} 42} \\ \hline
\multicolumn{1}{|c|}{{\color[HTML]{333333} \textbf{Rest (Averaged)}}} &
  \multicolumn{1}{c|}{{\color[HTML]{333333} 0.13043478}} &
  \multicolumn{1}{c|}{{\color[HTML]{333333} 0.91304348}} &
  \multicolumn{1}{c|}{{\color[HTML]{333333} 2.2173913}} &
  \multicolumn{1}{c|}{{\color[HTML]{333333} 9.34782609}} &
  \multicolumn{1}{c|}{{\color[HTML]{333333} -}} &
  \multicolumn{1}{c|}{{\color[HTML]{333333} -}} &
  \multicolumn{1}{c|}{{\color[HTML]{333333} -}} &
  {\color[HTML]{333333} -} \\ \hline
\end{tabular}%
}
\caption{Top $k$ occurrence for the Top 7 features of the DHI dataset as ranked by LIME during explanation of 101 test cases for model = Random Forest}
\label{tab:lime-rf-rank}
\end{table}

\begin{table}[H]
\centering
\resizebox{\columnwidth}{!}{%
\begin{tabular}{|ccccccccc|}
\hline
\multicolumn{9}{|c|}{{\color[HTML]{333333} \textbf{Top k-th Occurrences of high level features in LIME explanations (Model= Voting Classifier(1:5:1))}}} \\ \hline
\multicolumn{1}{|c|}{{\color[HTML]{333333} }} &
  \multicolumn{4}{c|}{{\color[HTML]{333333} \textit{\textbf{For All Features}}}} &
  \multicolumn{4}{c|}{{\color[HTML]{333333} \textit{\textbf{For Top 7 Features}}}} \\ \cline{2-9} 
\multicolumn{1}{|c|}{\multirow{-2}{*}{{\color[HTML]{333333} \textbf{\begin{tabular}[c]{@{}c@{}}Feature\\ Name\end{tabular}}}}} &
  \multicolumn{1}{c|}{{\color[HTML]{333333} \textit{\textbf{\begin{tabular}[c]{@{}c@{}}Top 1\\ occurrence\end{tabular}}}}} &
  \multicolumn{1}{c|}{{\color[HTML]{333333} \textit{\textbf{\begin{tabular}[c]{@{}c@{}}Top 2\\ occurrence\end{tabular}}}}} &
  \multicolumn{1}{c|}{{\color[HTML]{333333} \textit{\textbf{\begin{tabular}[c]{@{}c@{}}Top 3\\ occurrence\end{tabular}}}}} &
  \multicolumn{1}{c|}{{\color[HTML]{333333} \textit{\textbf{\begin{tabular}[c]{@{}c@{}}Top 5\\ occurrence\end{tabular}}}}} &
  \multicolumn{1}{c|}{{\color[HTML]{333333} \textit{\textbf{\begin{tabular}[c]{@{}c@{}}Top 1\\ occurrence\end{tabular}}}}} &
  \multicolumn{1}{c|}{{\color[HTML]{333333} \textit{\textbf{\begin{tabular}[c]{@{}c@{}}Top 2\\ occurrence\end{tabular}}}}} &
  \multicolumn{1}{c|}{{\color[HTML]{333333} \textit{\textbf{\begin{tabular}[c]{@{}c@{}}Top 3\\ occurrence\end{tabular}}}}} &
  {\color[HTML]{333333} \textit{\textbf{\begin{tabular}[c]{@{}c@{}}Top 5\\ occurrence\end{tabular}}}} \\ \hline
\multicolumn{1}{|c|}{{\color[HTML]{333333} \textbf{Feature 1}}} &
  \multicolumn{1}{c|}{{\color[HTML]{333333} 15}} &
  \multicolumn{1}{c|}{{\color[HTML]{333333} 28}} &
  \multicolumn{1}{c|}{{\color[HTML]{333333} 55}} &
  \multicolumn{1}{c|}{{\color[HTML]{333333} 83}} &
  \multicolumn{1}{c|}{{\color[HTML]{333333} 2}} &
  \multicolumn{1}{c|}{{\color[HTML]{333333} 22}} &
  \multicolumn{1}{c|}{{\color[HTML]{333333} 68}} &
  {\color[HTML]{333333} 101} \\ \hline
\multicolumn{1}{|c|}{{\color[HTML]{333333} \textbf{Feature 2}}} &
  \multicolumn{1}{c|}{{\color[HTML]{333333} 12}} &
  \multicolumn{1}{c|}{{\color[HTML]{333333} 17}} &
  \multicolumn{1}{c|}{{\color[HTML]{333333} 38}} &
  \multicolumn{1}{c|}{{\color[HTML]{333333} 54}} &
  \multicolumn{1}{c|}{{\color[HTML]{333333} 11}} &
  \multicolumn{1}{c|}{{\color[HTML]{333333} 22}} &
  \multicolumn{1}{c|}{{\color[HTML]{333333} 56}} &
  {\color[HTML]{333333} 93} \\ \hline
\multicolumn{1}{|c|}{{\color[HTML]{333333} \textbf{\begin{tabular}[c]{@{}c@{}}Feature 3\end{tabular}}}} &
  \multicolumn{1}{c|}{{\color[HTML]{CB0000} \textbf{8}}} &
  \multicolumn{1}{c|}{{\color[HTML]{CB0000} \textbf{52}}} &
  \multicolumn{1}{c|}{{\color[HTML]{333333} 55}} &
  \multicolumn{1}{c|}{{\color[HTML]{333333} 76}} &
  \multicolumn{1}{c|}{{\color[HTML]{CB0000} \textbf{5}}} &
  \multicolumn{1}{c|}{{\color[HTML]{CB0000} \textbf{65}}} &
  \multicolumn{1}{c|}{{\color[HTML]{333333} 73}} &
  {\color[HTML]{333333} 99} \\ \hline
\multicolumn{1}{|c|}{{\color[HTML]{333333} \textbf{Feature 4}}} &
  \multicolumn{1}{c|}{{\color[HTML]{333333} 60}} &
  \multicolumn{1}{c|}{{\color[HTML]{333333} 83}} &
  \multicolumn{1}{c|}{{\color[HTML]{333333} 92}} &
  \multicolumn{1}{c|}{{\color[HTML]{333333} 100}} &
  \multicolumn{1}{c|}{{\color[HTML]{333333} 83}} &
  \multicolumn{1}{c|}{{\color[HTML]{333333} 92}} &
  \multicolumn{1}{c|}{{\color[HTML]{333333} 100}} &
  {\color[HTML]{333333} 101} \\ \hline
\multicolumn{1}{|c|}{{\color[HTML]{333333} \textbf{\begin{tabular}[c]{@{}c@{}}Feature 5\end{tabular}}}} &
  \multicolumn{1}{c|}{{\color[HTML]{333333} 0}} &
  \multicolumn{1}{c|}{{\color[HTML]{333333} 0}} &
  \multicolumn{1}{c|}{{\color[HTML]{333333} 0}} &
  \multicolumn{1}{c|}{{\color[HTML]{333333} 0}} &
  \multicolumn{1}{c|}{{\color[HTML]{333333} 0}} &
  \multicolumn{1}{c|}{{\color[HTML]{333333} 0}} &
  \multicolumn{1}{c|}{{\color[HTML]{333333} 0}} &
  {\color[HTML]{333333} 21} \\ \hline
\multicolumn{1}{|c|}{{\color[HTML]{333333} \textbf{\begin{tabular}[c]{@{}c@{}}Feature 6\end{tabular}}}} &
  \multicolumn{1}{c|}{{\color[HTML]{333333} 0}} &
  \multicolumn{1}{c|}{{\color[HTML]{333333} 0}} &
  \multicolumn{1}{c|}{{\color[HTML]{333333} 0}} &
  \multicolumn{1}{c|}{{\color[HTML]{333333} 0}} &
  \multicolumn{1}{c|}{{\color[HTML]{333333} 0}} &
  \multicolumn{1}{c|}{{\color[HTML]{333333} 0}} &
  \multicolumn{1}{c|}{{\color[HTML]{333333} 0}} &
  {\color[HTML]{333333} 10} \\ \hline
\multicolumn{1}{|c|}{{\color[HTML]{333333} \textbf{\begin{tabular}[c]{@{}c@{}}Feature 7\end{tabular}}}} &
  \multicolumn{1}{c|}{{\color[HTML]{333333} 2}} &
  \multicolumn{1}{c|}{{\color[HTML]{333333} 5}} &
  \multicolumn{1}{c|}{{\color[HTML]{333333} 14}} &
  \multicolumn{1}{c|}{{\color[HTML]{333333} 21}} &
  \multicolumn{1}{c|}{{\color[HTML]{333333} 0}} &
  \multicolumn{1}{c|}{{\color[HTML]{333333} 1}} &
  \multicolumn{1}{c|}{{\color[HTML]{333333} 6}} &
  {\color[HTML]{333333} 80} \\ \hline
\multicolumn{1}{|c|}{{\color[HTML]{333333} \textbf{Rest (Averaged)}}} &
  \multicolumn{1}{c|}{{\color[HTML]{333333} 0.17391304}} &
  \multicolumn{1}{c|}{{\color[HTML]{333333} 0.73913043}} &
  \multicolumn{1}{c|}{{\color[HTML]{333333} 2.13043478}} &
  \multicolumn{1}{c|}{{\color[HTML]{333333} 7.43478261}} &
  \multicolumn{1}{c|}{{\color[HTML]{333333} -}} &
  \multicolumn{1}{c|}{{\color[HTML]{333333} -}} &
  \multicolumn{1}{c|}{{\color[HTML]{333333} -}} &
  {\color[HTML]{333333} -} \\ \hline
\end{tabular}%
}
\caption{Top $k$ occurrence for the top seven features of the DHI dataset as ranked by LIME during explanation of 101 test cases for model = Voting Classifier (1:5:1)}
\label{tab:lime-vc-rank}
\end{table}

\subsubsection{Top $k$ occurrence for SHAP Explanations:}

\begin{table}[H]
\centering
\resizebox{\columnwidth}{!}{%
\begin{tabular}{|ccccccccc|}
\hline
\multicolumn{9}{|c|}{{\color[HTML]{333333} \textbf{Top k-th Occurrences of high level features in SHAP explanations (Model= Logistic Regression)}}} \\ \hline
\multicolumn{1}{|c|}{{\color[HTML]{333333} }} &
  \multicolumn{4}{c|}{{\color[HTML]{333333} \textit{\textbf{For All Features}}}} &
  \multicolumn{4}{c|}{{\color[HTML]{333333} \textit{\textbf{For Top 7 Features}}}} \\ \cline{2-9} 
\multicolumn{1}{|c|}{\multirow{-2}{*}{{\color[HTML]{333333} \textbf{\begin{tabular}[c]{@{}c@{}}Feature\\ Name\end{tabular}}}}} &
  \multicolumn{1}{c|}{{\color[HTML]{333333} \textit{\textbf{\begin{tabular}[c]{@{}c@{}}Top 1\\ occurrence\end{tabular}}}}} &
  \multicolumn{1}{c|}{{\color[HTML]{333333} \textit{\textbf{\begin{tabular}[c]{@{}c@{}}Top 2\\ occurrence\end{tabular}}}}} &
  \multicolumn{1}{c|}{{\color[HTML]{333333} \textit{\textbf{\begin{tabular}[c]{@{}c@{}}Top 3\\ occurrence\end{tabular}}}}} &
  \multicolumn{1}{c|}{{\color[HTML]{333333} \textit{\textbf{\begin{tabular}[c]{@{}c@{}}Top 5\\ occurrence\end{tabular}}}}} &
  \multicolumn{1}{c|}{{\color[HTML]{333333} \textit{\textbf{\begin{tabular}[c]{@{}c@{}}Top 1\\ occurrence\end{tabular}}}}} &
  \multicolumn{1}{c|}{{\color[HTML]{333333} \textit{\textbf{\begin{tabular}[c]{@{}c@{}}Top 2\\ occurrence\end{tabular}}}}} &
  \multicolumn{1}{c|}{{\color[HTML]{333333} \textit{\textbf{\begin{tabular}[c]{@{}c@{}}Top 3\\ occurrence\end{tabular}}}}} &
  {\color[HTML]{333333} \textit{\textbf{\begin{tabular}[c]{@{}c@{}}Top 5\\ occurrence\end{tabular}}}} \\ \hline
\multicolumn{1}{|c|}{{\color[HTML]{333333} \textbf{Feature 1}}} &
  \multicolumn{1}{c|}{{\color[HTML]{333333} 0}} &
  \multicolumn{1}{c|}{{\color[HTML]{333333} 0}} &
  \multicolumn{1}{c|}{{\color[HTML]{333333} 0}} &
  \multicolumn{1}{c|}{{\color[HTML]{333333} 0}} &
  \multicolumn{1}{c|}{{\color[HTML]{333333} 3}} &
  \multicolumn{1}{c|}{{\color[HTML]{333333} 38}} &
  \multicolumn{1}{c|}{{\color[HTML]{333333} 56}} &
  {\color[HTML]{333333} 96} \\ \hline
\multicolumn{1}{|c|}{{\color[HTML]{333333} \textbf{Feature 2}}} &
  \multicolumn{1}{c|}{{\color[HTML]{333333} 0}} &
  \multicolumn{1}{c|}{{\color[HTML]{333333} 0}} &
  \multicolumn{1}{c|}{{\color[HTML]{333333} 1}} &
  \multicolumn{1}{c|}{{\color[HTML]{333333} 6}} &
  \multicolumn{1}{c|}{{\color[HTML]{333333} 8}} &
  \multicolumn{1}{c|}{{\color[HTML]{333333} 45}} &
  \multicolumn{1}{c|}{{\color[HTML]{333333} 70}} &
  {\color[HTML]{333333} 85} \\ \hline
\multicolumn{1}{|c|}{{\color[HTML]{333333} \textbf{\begin{tabular}[c]{@{}c@{}}Feature 3\end{tabular}}}} &
  \multicolumn{1}{c|}{{\color[HTML]{333333} 0}} &
  \multicolumn{1}{c|}{{\color[HTML]{333333} 0}} &
  \multicolumn{1}{c|}{{\color[HTML]{333333} 0}} &
  \multicolumn{1}{c|}{{\color[HTML]{333333} 0}} &
  \multicolumn{1}{c|}{{\color[HTML]{333333} 0}} &
  \multicolumn{1}{c|}{{\color[HTML]{333333} 18}} &
  \multicolumn{1}{c|}{{\color[HTML]{333333} 48}} &
  {\color[HTML]{333333} 78} \\ \hline
\multicolumn{1}{|c|}{{\color[HTML]{333333} \textbf{Feature 4}}} &
  \multicolumn{1}{c|}{{\color[HTML]{333333} 56}} &
  \multicolumn{1}{c|}{{\color[HTML]{333333} 67}} &
  \multicolumn{1}{c|}{{\color[HTML]{333333} 72}} &
  \multicolumn{1}{c|}{{\color[HTML]{333333} 84}} &
  \multicolumn{1}{c|}{{\color[HTML]{333333} 90}} &
  \multicolumn{1}{c|}{{\color[HTML]{333333} 92}} &
  \multicolumn{1}{c|}{{\color[HTML]{333333} 95}} &
  {\color[HTML]{333333} 97} \\ \hline
\multicolumn{1}{|c|}{{\color[HTML]{333333} \textbf{\begin{tabular}[c]{@{}c@{}}Feature 5\end{tabular}}}} &
  \multicolumn{1}{c|}{{\color[HTML]{333333} 0}} &
  \multicolumn{1}{c|}{{\color[HTML]{333333} 0}} &
  \multicolumn{1}{c|}{{\color[HTML]{333333} 0}} &
  \multicolumn{1}{c|}{{\color[HTML]{333333} 0}} &
  \multicolumn{1}{c|}{{\color[HTML]{333333} 0}} &
  \multicolumn{1}{c|}{{\color[HTML]{333333} 0}} &
  \multicolumn{1}{c|}{{\color[HTML]{333333} 0}} &
  {\color[HTML]{333333} 15} \\ \hline
\multicolumn{1}{|c|}{{\color[HTML]{333333} \textbf{\begin{tabular}[c]{@{}c@{}}Feature 6\end{tabular}}}} &
  \multicolumn{1}{c|}{{\color[HTML]{333333} 0}} &
  \multicolumn{1}{c|}{{\color[HTML]{333333} 0}} &
  \multicolumn{1}{c|}{{\color[HTML]{333333} 0}} &
  \multicolumn{1}{c|}{{\color[HTML]{333333} 0}} &
  \multicolumn{1}{c|}{{\color[HTML]{333333} 0}} &
  \multicolumn{1}{c|}{{\color[HTML]{333333} 9}} &
  \multicolumn{1}{c|}{{\color[HTML]{333333} 22}} &
  {\color[HTML]{333333} 70} \\ \hline
\multicolumn{1}{|c|}{{\color[HTML]{333333} \textbf{\begin{tabular}[c]{@{}c@{}}Feature 7\end{tabular}}}} &
  \multicolumn{1}{c|}{{\color[HTML]{333333} 0}} &
  \multicolumn{1}{c|}{{\color[HTML]{333333} 0}} &
  \multicolumn{1}{c|}{{\color[HTML]{333333} 0}} &
  \multicolumn{1}{c|}{{\color[HTML]{333333} 0}} &
  \multicolumn{1}{c|}{{\color[HTML]{333333} 0}} &
  \multicolumn{1}{c|}{{\color[HTML]{333333} 0}} &
  \multicolumn{1}{c|}{{\color[HTML]{333333} 12}} &
  {\color[HTML]{333333} 64} \\ \hline
\multicolumn{1}{|c|}{{\color[HTML]{333333} \textbf{Rest (Averaged)}}} &
  \multicolumn{1}{c|}{{\color[HTML]{333333} 1.956521739}} &
  \multicolumn{1}{c|}{{\color[HTML]{333333} 5.869565217}} &
  \multicolumn{1}{c|}{{\color[HTML]{333333} \textbf{10}}} &
  \multicolumn{1}{c|}{{\color[HTML]{333333} \textbf{18.04347826}}} &
  \multicolumn{1}{c|}{{\color[HTML]{333333} -}} &
  \multicolumn{1}{c|}{{\color[HTML]{333333} -}} &
  \multicolumn{1}{c|}{{\color[HTML]{333333} -}} &
  {\color[HTML]{333333} -} \\ \hline
\end{tabular}%
}
\caption{Top $k$ occurrence for the Top 7 features of the DHI dataset as ranked by SHAP during explanation of 101 test cases for model = Logistic Regression}
\label{tab:shap-lr-rank}
\end{table}

\begin{table}[H]
\centering
\resizebox{\columnwidth}{!}{%
\begin{tabular}{|ccccccccc|}
\hline
\multicolumn{9}{|c|}{{\color[HTML]{333333} \textbf{Top k-th Occurrences of high level features in SHAP explanations (Model= Gaussian NB)}}} \\ \hline
\multicolumn{1}{|c|}{{\color[HTML]{333333} }} &
  \multicolumn{4}{c|}{{\color[HTML]{333333} \textit{\textbf{For All Features}}}} &
  \multicolumn{4}{c|}{{\color[HTML]{333333} \textit{\textbf{For Top 7 Features}}}} \\ \cline{2-9} 
\multicolumn{1}{|c|}{\multirow{-2}{*}{{\color[HTML]{333333} \textbf{\begin{tabular}[c]{@{}c@{}}Feature\\ Name\end{tabular}}}}} &
  \multicolumn{1}{c|}{{\color[HTML]{333333} \textit{\textbf{\begin{tabular}[c]{@{}c@{}}Top 1\\ occurrence\end{tabular}}}}} &
  \multicolumn{1}{c|}{{\color[HTML]{333333} \textit{\textbf{\begin{tabular}[c]{@{}c@{}}Top 2\\ occurrence\end{tabular}}}}} &
  \multicolumn{1}{c|}{{\color[HTML]{333333} \textit{\textbf{\begin{tabular}[c]{@{}c@{}}Top 3\\ occurrence\end{tabular}}}}} &
  \multicolumn{1}{c|}{{\color[HTML]{333333} \textit{\textbf{\begin{tabular}[c]{@{}c@{}}Top 5\\ occurrence\end{tabular}}}}} &
  \multicolumn{1}{c|}{{\color[HTML]{333333} \textit{\textbf{\begin{tabular}[c]{@{}c@{}}Top 1\\ occurrence\end{tabular}}}}} &
  \multicolumn{1}{c|}{{\color[HTML]{333333} \textit{\textbf{\begin{tabular}[c]{@{}c@{}}Top 2\\ occurrence\end{tabular}}}}} &
  \multicolumn{1}{c|}{{\color[HTML]{333333} \textit{\textbf{\begin{tabular}[c]{@{}c@{}}Top 3\\ occurrence\end{tabular}}}}} &
  {\color[HTML]{333333} \textit{\textbf{\begin{tabular}[c]{@{}c@{}}Top 5\\ occurrence\end{tabular}}}} \\ \hline
\multicolumn{1}{|c|}{{\color[HTML]{333333} \textbf{Feature 1}}} &
  \multicolumn{1}{c|}{{\color[HTML]{333333} 3}} &
  \multicolumn{1}{c|}{{\color[HTML]{333333} 6}} &
  \multicolumn{1}{c|}{{\color[HTML]{333333} 8}} &
  \multicolumn{1}{c|}{{\color[HTML]{333333} 27}} &
  \multicolumn{1}{c|}{{\color[HTML]{333333} 1}} &
  \multicolumn{1}{c|}{{\color[HTML]{333333} 9}} &
  \multicolumn{1}{c|}{{\color[HTML]{333333} 36}} &
  {\color[HTML]{333333} 95} \\ \hline
\multicolumn{1}{|c|}{{\color[HTML]{333333} \textbf{Feature 2}}} &
  \multicolumn{1}{c|}{{\color[HTML]{333333} 8}} &
  \multicolumn{1}{c|}{{\color[HTML]{333333} 17}} &
  \multicolumn{1}{c|}{{\color[HTML]{333333} 50}} &
  \multicolumn{1}{c|}{{\color[HTML]{333333} 71}} &
  \multicolumn{1}{c|}{{\color[HTML]{333333} 17}} &
  \multicolumn{1}{c|}{{\color[HTML]{333333} 31}} &
  \multicolumn{1}{c|}{{\color[HTML]{333333} 78}} &
  {\color[HTML]{333333} 98} \\ \hline
\multicolumn{1}{|c|}{{\color[HTML]{333333} \textbf{\begin{tabular}[c]{@{}c@{}}Feature 3\end{tabular}}}} &
  \multicolumn{1}{c|}{{\color[HTML]{CB0000} \textbf{3}}} &
  \multicolumn{1}{c|}{{\color[HTML]{CB0000} \textbf{53}}} &
  \multicolumn{1}{c|}{{\color[HTML]{333333} 75}} &
  \multicolumn{1}{c|}{{\color[HTML]{333333} 84}} &
  \multicolumn{1}{c|}{{\color[HTML]{CB0000} \textbf{7}}} &
  \multicolumn{1}{c|}{{\color[HTML]{CB0000} \textbf{69}}} &
  \multicolumn{1}{c|}{{\color[HTML]{333333} 87}} &
  {\color[HTML]{333333} 99} \\ \hline
\multicolumn{1}{|c|}{{\color[HTML]{333333} \textbf{Feature 4}}} &
  \multicolumn{1}{c|}{{\color[HTML]{333333} 71}} &
  \multicolumn{1}{c|}{{\color[HTML]{333333} 86}} &
  \multicolumn{1}{c|}{{\color[HTML]{333333} 88}} &
  \multicolumn{1}{c|}{{\color[HTML]{333333} 91}} &
  \multicolumn{1}{c|}{{\color[HTML]{333333} 74}} &
  \multicolumn{1}{c|}{{\color[HTML]{333333} 89}} &
  \multicolumn{1}{c|}{{\color[HTML]{333333} 95}} &
  {\color[HTML]{333333} 100} \\ \hline
\multicolumn{1}{|c|}{{\color[HTML]{333333} \textbf{\begin{tabular}[c]{@{}c@{}}Feature 5\end{tabular}}}} &
  \multicolumn{1}{c|}{{\color[HTML]{333333} 1}} &
  \multicolumn{1}{c|}{{\color[HTML]{333333} 1}} &
  \multicolumn{1}{c|}{{\color[HTML]{333333} 3}} &
  \multicolumn{1}{c|}{{\color[HTML]{333333} 9}} &
  \multicolumn{1}{c|}{{\color[HTML]{333333} 0}} &
  \multicolumn{1}{c|}{{\color[HTML]{333333} 0}} &
  \multicolumn{1}{c|}{{\color[HTML]{333333} 0}} &
  {\color[HTML]{333333} 26} \\ \hline
\multicolumn{1}{|c|}{{\color[HTML]{333333} \textbf{\begin{tabular}[c]{@{}c@{}}Feature 6\end{tabular}}}} &
  \multicolumn{1}{c|}{{\color[HTML]{333333} 0}} &
  \multicolumn{1}{c|}{{\color[HTML]{333333} 0}} &
  \multicolumn{1}{c|}{{\color[HTML]{333333} 0}} &
  \multicolumn{1}{c|}{{\color[HTML]{333333} 1}} &
  \multicolumn{1}{c|}{{\color[HTML]{333333} 0}} &
  \multicolumn{1}{c|}{{\color[HTML]{333333} 0}} &
  \multicolumn{1}{c|}{{\color[HTML]{333333} 0}} &
  {\color[HTML]{333333} 1} \\ \hline
\multicolumn{1}{|c|}{{\color[HTML]{333333} \textbf{\begin{tabular}[c]{@{}c@{}}Feature 7\end{tabular}}}} &
  \multicolumn{1}{c|}{{\color[HTML]{333333} 1}} &
  \multicolumn{1}{c|}{{\color[HTML]{333333} 3}} &
  \multicolumn{1}{c|}{{\color[HTML]{333333} 6}} &
  \multicolumn{1}{c|}{{\color[HTML]{333333} 20}} &
  \multicolumn{1}{c|}{{\color[HTML]{333333} 2}} &
  \multicolumn{1}{c|}{{\color[HTML]{333333} 4}} &
  \multicolumn{1}{c|}{{\color[HTML]{333333} 7}} &
  {\color[HTML]{333333} 86} \\ \hline
\multicolumn{1}{|c|}{{\color[HTML]{333333} \textbf{Rest (Averaged)}}} &
  \multicolumn{1}{c|}{{\color[HTML]{333333} 0.60869565}} &
  \multicolumn{1}{c|}{{\color[HTML]{333333} 1.56521739}} &
  \multicolumn{1}{c|}{{\color[HTML]{333333} 3.17391304}} &
  \multicolumn{1}{c|}{{\color[HTML]{333333} 8.7826087}} &
  \multicolumn{1}{c|}{{\color[HTML]{333333} -}} &
  \multicolumn{1}{c|}{{\color[HTML]{333333} -}} &
  \multicolumn{1}{c|}{{\color[HTML]{333333} -}} &
  {\color[HTML]{333333} -} \\ \hline
\end{tabular}%
}
\caption{Top $k$ occurrence for the top seven features of the DHI dataset as ranked by SHAP during explanation of 101 test cases for model = Gaussian NB}
\label{tab:shap-GNB-rank}
\end{table}

\begin{table}[H]
\centering
\resizebox{\columnwidth}{!}{%
\begin{tabular}{|ccccccccc|}
\hline
\multicolumn{9}{|c|}{{\color[HTML]{333333} \textbf{Top k-th Occurrences of high level features in SHAP explanations (Model= Random Forest)}}} \\ \hline
\multicolumn{1}{|c|}{{\color[HTML]{333333} }} &
  \multicolumn{4}{c|}{{\color[HTML]{333333} \textit{\textbf{For All Features}}}} &
  \multicolumn{4}{c|}{{\color[HTML]{333333} \textit{\textbf{For Top 7 Features}}}} \\ \cline{2-9} 
\multicolumn{1}{|c|}{\multirow{-2}{*}{{\color[HTML]{333333} \textbf{\begin{tabular}[c]{@{}c@{}}Feature\\ Name\end{tabular}}}}} &
  \multicolumn{1}{c|}{{\color[HTML]{333333} \textit{\textbf{\begin{tabular}[c]{@{}c@{}}Top 1\\ occurrence\end{tabular}}}}} &
  \multicolumn{1}{c|}{{\color[HTML]{333333} \textit{\textbf{\begin{tabular}[c]{@{}c@{}}Top 2\\ occurrence\end{tabular}}}}} &
  \multicolumn{1}{c|}{{\color[HTML]{333333} \textit{\textbf{\begin{tabular}[c]{@{}c@{}}Top 3\\ occurrence\end{tabular}}}}} &
  \multicolumn{1}{c|}{{\color[HTML]{333333} \textit{\textbf{\begin{tabular}[c]{@{}c@{}}Top 5\\ occurrence\end{tabular}}}}} &
  \multicolumn{1}{c|}{{\color[HTML]{333333} \textit{\textbf{\begin{tabular}[c]{@{}c@{}}Top 1\\ occurrence\end{tabular}}}}} &
  \multicolumn{1}{c|}{{\color[HTML]{333333} \textit{\textbf{\begin{tabular}[c]{@{}c@{}}Top 2\\ occurrence\end{tabular}}}}} &
  \multicolumn{1}{c|}{{\color[HTML]{333333} \textit{\textbf{\begin{tabular}[c]{@{}c@{}}Top 3\\ occurrence\end{tabular}}}}} &
  {\color[HTML]{333333} \textit{\textbf{\begin{tabular}[c]{@{}c@{}}Top 5\\ occurrence\end{tabular}}}} \\ \hline
\multicolumn{1}{|c|}{{\color[HTML]{333333} \textbf{Feature 1}}} &
  \multicolumn{1}{c|}{{\color[HTML]{333333} 0}} &
  \multicolumn{1}{c|}{{\color[HTML]{333333} 0}} &
  \multicolumn{1}{c|}{{\color[HTML]{333333} 0}} &
  \multicolumn{1}{c|}{{\color[HTML]{333333} 0}} &
  \multicolumn{1}{c|}{{\color[HTML]{333333} 0}} &
  \multicolumn{1}{c|}{{\color[HTML]{333333} 1}} &
  \multicolumn{1}{c|}{{\color[HTML]{333333} 7}} &
  {\color[HTML]{333333} 25} \\ \hline
\multicolumn{1}{|c|}{{\color[HTML]{333333} \textbf{Feature 2}}} &
  \multicolumn{1}{c|}{{\color[HTML]{CB0000} \textbf{19}}} &
  \multicolumn{1}{c|}{{\color[HTML]{CB0000} \textbf{34}}} &
  \multicolumn{1}{c|}{{\color[HTML]{333333} 68}} &
  \multicolumn{1}{c|}{{\color[HTML]{333333} 80}} &
  \multicolumn{1}{c|}{{\color[HTML]{333333} 11}} &
  \multicolumn{1}{c|}{{\color[HTML]{333333} 25}} &
  \multicolumn{1}{c|}{{\color[HTML]{333333} 74}} &
  {\color[HTML]{333333} 92} \\ \hline
\multicolumn{1}{|c|}{{\color[HTML]{333333} \textbf{\begin{tabular}[c]{@{}c@{}}Feature 3\end{tabular}}}} &
  \multicolumn{1}{c|}{{\color[HTML]{CB0000} \textbf{20}}} &
  \multicolumn{1}{c|}{{\color[HTML]{CB0000} \textbf{63}}} &
  \multicolumn{1}{c|}{{\color[HTML]{333333} 79}} &
  \multicolumn{1}{c|}{{\color[HTML]{333333} 85}} &
  \multicolumn{1}{c|}{{\color[HTML]{333333} 14}} &
  \multicolumn{1}{c|}{{\color[HTML]{333333} 78}} &
  \multicolumn{1}{c|}{{\color[HTML]{333333} 86}} &
  {\color[HTML]{333333} 98} \\ \hline
\multicolumn{1}{|c|}{{\color[HTML]{333333} \textbf{Feature 4}}} &
  \multicolumn{1}{c|}{{\color[HTML]{CB0000} \textbf{55}}} &
  \multicolumn{1}{c|}{{\color[HTML]{CB0000} \textbf{74}}} &
  \multicolumn{1}{c|}{{\color[HTML]{333333} 83}} &
  \multicolumn{1}{c|}{{\color[HTML]{333333} 89}} &
  \multicolumn{1}{c|}{{\color[HTML]{333333} 71}} &
  \multicolumn{1}{c|}{{\color[HTML]{333333} 85}} &
  \multicolumn{1}{c|}{{\color[HTML]{333333} 91}} &
  {\color[HTML]{333333} 98} \\ \hline
\multicolumn{1}{|c|}{{\color[HTML]{333333} \textbf{\begin{tabular}[c]{@{}c@{}}Feature 5\end{tabular}}}} &
  \multicolumn{1}{c|}{{\color[HTML]{333333} 1}} &
  \multicolumn{1}{c|}{{\color[HTML]{333333} 4}} &
  \multicolumn{1}{c|}{{\color[HTML]{333333} 4}} &
  \multicolumn{1}{c|}{{\color[HTML]{333333} 14}} &
  \multicolumn{1}{c|}{{\color[HTML]{333333} 1}} &
  \multicolumn{1}{c|}{{\color[HTML]{CB0000} \textbf{2}}} &
  \multicolumn{1}{c|}{{\color[HTML]{CB0000} \textbf{12}}} &
  {\color[HTML]{333333} 60} \\ \hline
\multicolumn{1}{|c|}{{\color[HTML]{333333} \textbf{\begin{tabular}[c]{@{}c@{}}Feature 6\end{tabular}}}} &
  \multicolumn{1}{c|}{{\color[HTML]{333333} 0}} &
  \multicolumn{1}{c|}{{\color[HTML]{333333} 1}} &
  \multicolumn{1}{c|}{{\color[HTML]{333333} 2}} &
  \multicolumn{1}{c|}{{\color[HTML]{333333} 8}} &
  \multicolumn{1}{c|}{{\color[HTML]{333333} 3}} &
  \multicolumn{1}{c|}{{\color[HTML]{CB0000} \textbf{9}}} &
  \multicolumn{1}{c|}{{\color[HTML]{CB0000} \textbf{20}}} &
  {\color[HTML]{333333} 73} \\ \hline
\multicolumn{1}{|c|}{{\color[HTML]{333333} \textbf{\begin{tabular}[c]{@{}c@{}}Feature 7\end{tabular}}}} &
  \multicolumn{1}{c|}{{\color[HTML]{333333} 0}} &
  \multicolumn{1}{c|}{{\color[HTML]{333333} 1}} &
  \multicolumn{1}{c|}{{\color[HTML]{333333} 9}} &
  \multicolumn{1}{c|}{{\color[HTML]{333333} 20}} &
  \multicolumn{1}{c|}{{\color[HTML]{333333} 1}} &
  \multicolumn{1}{c|}{{\color[HTML]{CB0000} \textbf{2}}} &
  \multicolumn{1}{c|}{{\color[HTML]{CB0000} \textbf{13}}} &
  {\color[HTML]{333333} 59} \\ \hline
\multicolumn{1}{|c|}{{\color[HTML]{333333} \textbf{Rest (Averaged)}}} &
  \multicolumn{1}{c|}{{\color[HTML]{333333} 0.26086957}} &
  \multicolumn{1}{c|}{{\color[HTML]{333333} 1.08695652}} &
  \multicolumn{1}{c|}{{\color[HTML]{333333} 2.52173913}} &
  \multicolumn{1}{c|}{{\color[HTML]{333333} 9.08695652}} &
  \multicolumn{1}{c|}{{\color[HTML]{333333} -}} &
  \multicolumn{1}{c|}{{\color[HTML]{333333} -}} &
  \multicolumn{1}{c|}{{\color[HTML]{333333} -}} &
  {\color[HTML]{333333} -} \\ \hline
\end{tabular}%
}
\caption{Top $k$ occurrence for the Top 7 features of the DHI dataset as ranked by SHAP during explanation of 101 test cases for model = Random Forest}
\label{tab:shap-rf-rank}
\end{table}

\begin{table}[H]
\centering
\resizebox{\columnwidth}{!}{%
\begin{tabular}{|ccccccccc|}
\hline
\multicolumn{9}{|c|}{{\color[HTML]{333333} \textbf{Top k-th Occurrences of high level features in SHAP explanations (Model= Voting Classifier(1:5:1))}}} \\ \hline
\multicolumn{1}{|c|}{{\color[HTML]{333333} }} &
  \multicolumn{4}{c|}{{\color[HTML]{333333} \textit{\textbf{For All Features}}}} &
  \multicolumn{4}{c|}{{\color[HTML]{333333} \textit{\textbf{For Top 7 Features}}}} \\ \cline{2-9} 
\multicolumn{1}{|c|}{\multirow{-2}{*}{{\color[HTML]{333333} \textbf{\begin{tabular}[c]{@{}c@{}}Feature\\ Name\end{tabular}}}}} &
  \multicolumn{1}{c|}{{\color[HTML]{333333} \textit{\textbf{\begin{tabular}[c]{@{}c@{}}Top 1\\ occurrence\end{tabular}}}}} &
  \multicolumn{1}{c|}{{\color[HTML]{333333} \textit{\textbf{\begin{tabular}[c]{@{}c@{}}Top 2\\ occurrence\end{tabular}}}}} &
  \multicolumn{1}{c|}{{\color[HTML]{333333} \textit{\textbf{\begin{tabular}[c]{@{}c@{}}Top 3\\ occurrence\end{tabular}}}}} &
  \multicolumn{1}{c|}{{\color[HTML]{333333} \textit{\textbf{\begin{tabular}[c]{@{}c@{}}Top 5\\ occurrence\end{tabular}}}}} &
  \multicolumn{1}{c|}{{\color[HTML]{333333} \textit{\textbf{\begin{tabular}[c]{@{}c@{}}Top 1\\ occurrence\end{tabular}}}}} &
  \multicolumn{1}{c|}{{\color[HTML]{333333} \textit{\textbf{\begin{tabular}[c]{@{}c@{}}Top 2\\ occurrence\end{tabular}}}}} &
  \multicolumn{1}{c|}{{\color[HTML]{333333} \textit{\textbf{\begin{tabular}[c]{@{}c@{}}Top 3\\ occurrence\end{tabular}}}}} &
  {\color[HTML]{333333} \textit{\textbf{\begin{tabular}[c]{@{}c@{}}Top 5\\ occurrence\end{tabular}}}} \\ \hline
\multicolumn{1}{|c|}{{\color[HTML]{333333} \textbf{Feature 1}}} &
  \multicolumn{1}{c|}{{\color[HTML]{333333} 0}} &
  \multicolumn{1}{c|}{{\color[HTML]{333333} 2}} &
  \multicolumn{1}{c|}{{\color[HTML]{333333} 9}} &
  \multicolumn{1}{c|}{{\color[HTML]{333333} 15}} &
  \multicolumn{1}{c|}{{\color[HTML]{333333} 1}} &
  \multicolumn{1}{c|}{{\color[HTML]{333333} 2}} &
  \multicolumn{1}{c|}{{\color[HTML]{333333} 6}} &
  {\color[HTML]{333333} 25} \\ \hline
\multicolumn{1}{|c|}{{\color[HTML]{333333} \textbf{Feature 2}}} &
  \multicolumn{1}{c|}{{\color[HTML]{333333} 10}} &
  \multicolumn{1}{c|}{{\color[HTML]{333333} 21}} &
  \multicolumn{1}{c|}{{\color[HTML]{333333} 42}} &
  \multicolumn{1}{c|}{{\color[HTML]{333333} 65}} &
  \multicolumn{1}{c|}{{\color[HTML]{333333} 19}} &
  \multicolumn{1}{c|}{{\color[HTML]{333333} 28}} &
  \multicolumn{1}{c|}{{\color[HTML]{333333} 79}} &
  {\color[HTML]{333333} 96} \\ \hline
\multicolumn{1}{|c|}{{\color[HTML]{333333} \textbf{\begin{tabular}[c]{@{}c@{}}Feature 3\end{tabular}}}} &
  \multicolumn{1}{c|}{{\color[HTML]{CB0000} \textbf{2}}} &
  \multicolumn{1}{c|}{{\color[HTML]{CB0000} \textbf{62}}} &
  \multicolumn{1}{c|}{{\color[HTML]{333333} 77}} &
  \multicolumn{1}{c|}{{\color[HTML]{333333} 84}} &
  \multicolumn{1}{c|}{{\color[HTML]{CB0000} \textbf{1}}} &
  \multicolumn{1}{c|}{{\color[HTML]{CB0000} \textbf{73}}} &
  \multicolumn{1}{c|}{{\color[HTML]{333333} 88}} &
  {\color[HTML]{333333} 96} \\ \hline
\multicolumn{1}{|c|}{{\color[HTML]{333333} \textbf{Feature 4}}} &
  \multicolumn{1}{c|}{{\color[HTML]{333333} 76}} &
  \multicolumn{1}{c|}{{\color[HTML]{333333} 83}} &
  \multicolumn{1}{c|}{{\color[HTML]{333333} 87}} &
  \multicolumn{1}{c|}{{\color[HTML]{333333} 92}} &
  \multicolumn{1}{c|}{{\color[HTML]{333333} 79}} &
  \multicolumn{1}{c|}{{\color[HTML]{333333} 93}} &
  \multicolumn{1}{c|}{{\color[HTML]{333333} 98}} &
  {\color[HTML]{333333} 100} \\ \hline
\multicolumn{1}{|c|}{{\color[HTML]{333333} \textbf{\begin{tabular}[c]{@{}c@{}}Feature 5\end{tabular}}}} &
  \multicolumn{1}{c|}{{\color[HTML]{333333} 0}} &
  \multicolumn{1}{c|}{{\color[HTML]{333333} 0}} &
  \multicolumn{1}{c|}{{\color[HTML]{333333} 2}} &
  \multicolumn{1}{c|}{{\color[HTML]{333333} 7}} &
  \multicolumn{1}{c|}{{\color[HTML]{333333} 0}} &
  \multicolumn{1}{c|}{{\color[HTML]{333333} 0}} &
  \multicolumn{1}{c|}{{\color[HTML]{333333} 1}} &
  {\color[HTML]{333333} 60} \\ \hline
\multicolumn{1}{|c|}{{\color[HTML]{333333} \textbf{\begin{tabular}[c]{@{}c@{}}Feature 6\end{tabular}}}} &
  \multicolumn{1}{c|}{{\color[HTML]{333333} 0}} &
  \multicolumn{1}{c|}{{\color[HTML]{333333} 0}} &
  \multicolumn{1}{c|}{{\color[HTML]{333333} 0}} &
  \multicolumn{1}{c|}{{\color[HTML]{333333} 0}} &
  \multicolumn{1}{c|}{{\color[HTML]{333333} 0}} &
  \multicolumn{1}{c|}{{\color[HTML]{333333} 0}} &
  \multicolumn{1}{c|}{{\color[HTML]{333333} 3}} &
  {\color[HTML]{333333} 44} \\ \hline
\multicolumn{1}{|c|}{{\color[HTML]{333333} \textbf{\begin{tabular}[c]{@{}c@{}}Feature 7\end{tabular}}}} &
  \multicolumn{1}{c|}{{\color[HTML]{333333} 2}} &
  \multicolumn{1}{c|}{{\color[HTML]{333333} 4}} &
  \multicolumn{1}{c|}{{\color[HTML]{333333} 9}} &
  \multicolumn{1}{c|}{{\color[HTML]{333333} 30}} &
  \multicolumn{1}{c|}{{\color[HTML]{333333} 1}} &
  \multicolumn{1}{c|}{{\color[HTML]{333333} 6}} &
  \multicolumn{1}{c|}{{\color[HTML]{333333} 28}} &
  {\color[HTML]{333333} 84} \\ \hline
\multicolumn{1}{|c|}{{\color[HTML]{333333} \textbf{Rest (Averaged)}}} &
  \multicolumn{1}{c|}{{\color[HTML]{333333} 0.47826087}} &
  \multicolumn{1}{c|}{{\color[HTML]{333333} 1.30434783}} &
  \multicolumn{1}{c|}{{\color[HTML]{333333} 3.34782609}} &
  \multicolumn{1}{c|}{{\color[HTML]{333333} 9.2173913}} &
  \multicolumn{1}{c|}{{\color[HTML]{333333} -}} &
  \multicolumn{1}{c|}{{\color[HTML]{333333} -}} &
  \multicolumn{1}{c|}{{\color[HTML]{333333} -}} &
  {\color[HTML]{333333} -} \\ \hline
\end{tabular}%
}
\caption{Top $k$ occurrence for the top seven features of the DHI dataset as ranked by SHAP during explanation of 101 test cases for model = Voting Classifier (1:5:1)}
\label{tab:shap-vc-rank}
\end{table}








\clearpage
\bibliographystyle{seg}  
\bibliography{ref}
\newpage
\listoffigures
\newpage

\begin{figure*} 
  \centering
  \includegraphics[width=\textwidth,height=0.3\textheight]{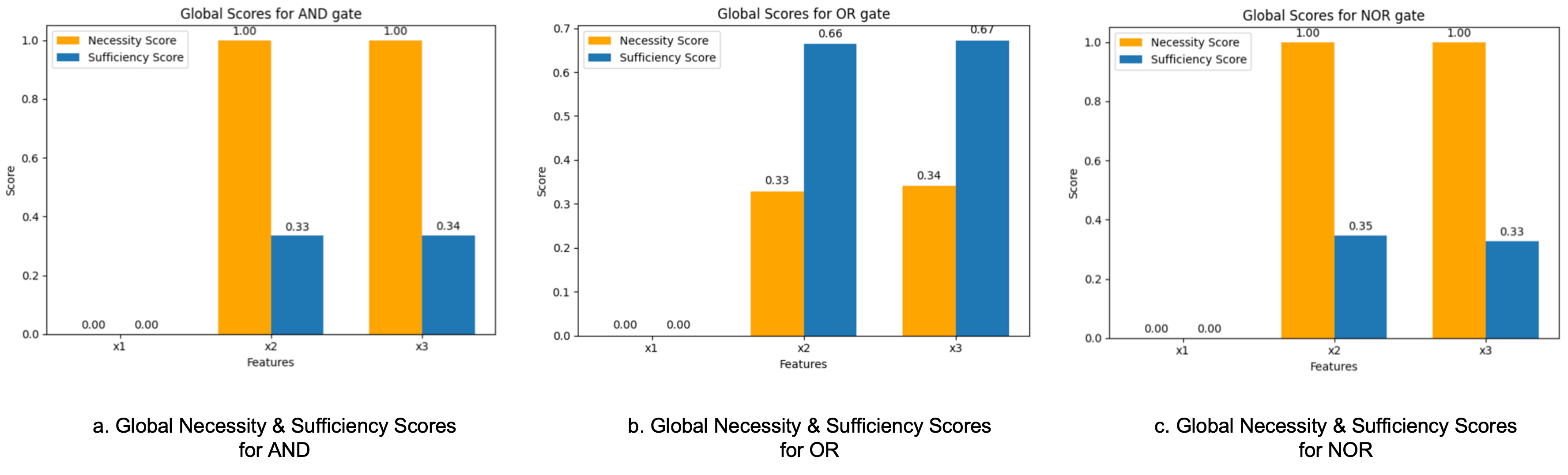} 

  \caption{Toy example to validate that the necessity and sufficiency scores generated by the selected method are synonymous with logically calculated impact score of a cause leading to an effect. }  \label{fig:TOY}

\end{figure*}

\begin{figure*} 
  \centering
  \includegraphics[width=\textwidth,height=0.55\textheight]{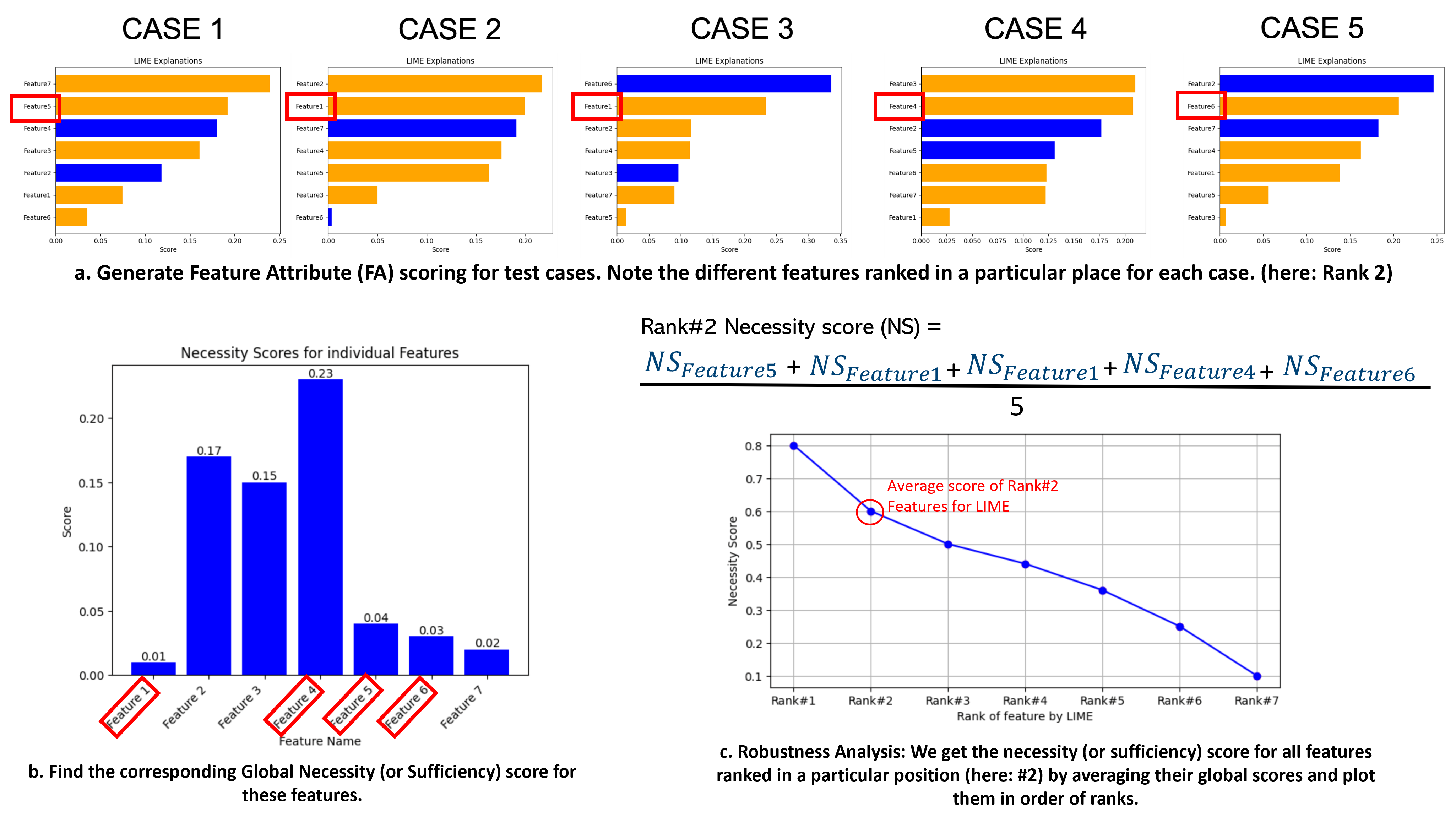} 
  \caption{Robustness Analysis of LIME scores: Calculating mean Necessity Score for Rank $2$. (a) Record the Feature names which occupy the same rank in all cases. (b) Match them with their corresponding Global Necessity Scores. (c) The average of the Global Scores of the features in a particular rank (here $2$) gives us the robustness analysis of LIME explanation of the decisions made by the model on the dataset. }  \label{fig:robust}

\end{figure*}

\begin{figure*}
  \centering
  \includegraphics[width=\textwidth]{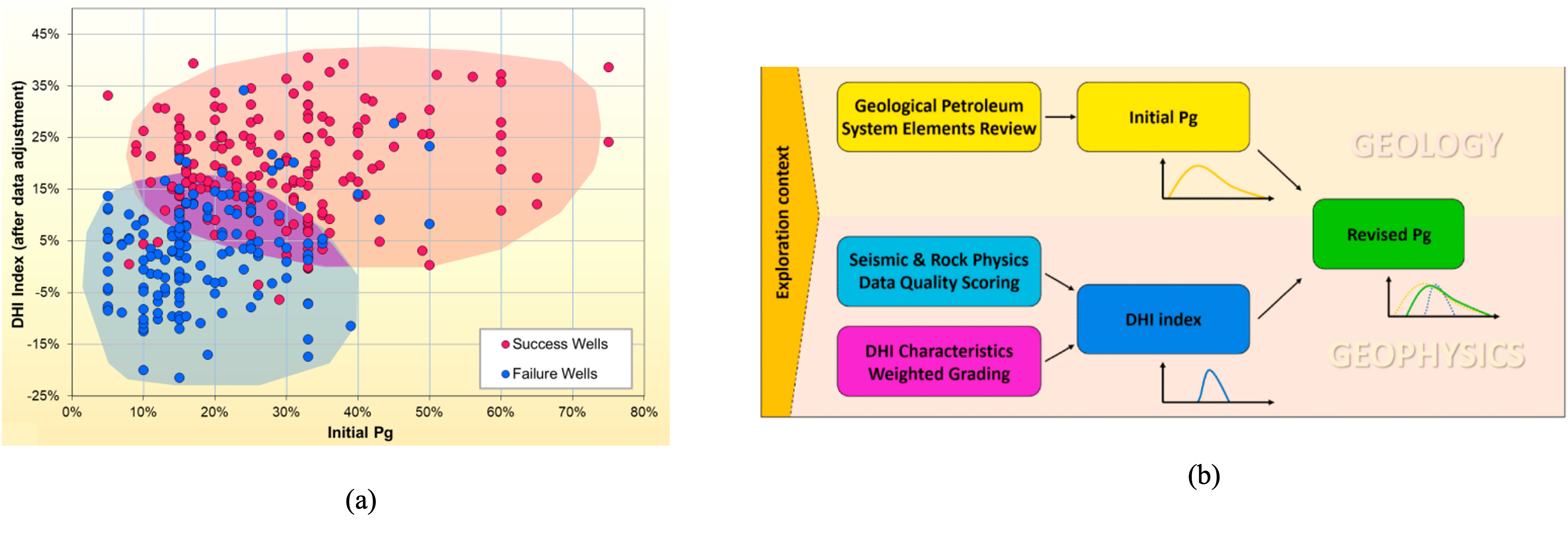}
  \caption{(a) A plot of the well outcomes by DHI Index vs. Initial Pg.  (b) The risk factors assessment flow chart: This figure presents the main steps of the Pg calculation for a given prospect.}
  \label{fig:dhi}
\end{figure*}

\begin{figure*} 
  \centering
  \includegraphics[width=\textwidth,height=0.55\textheight]{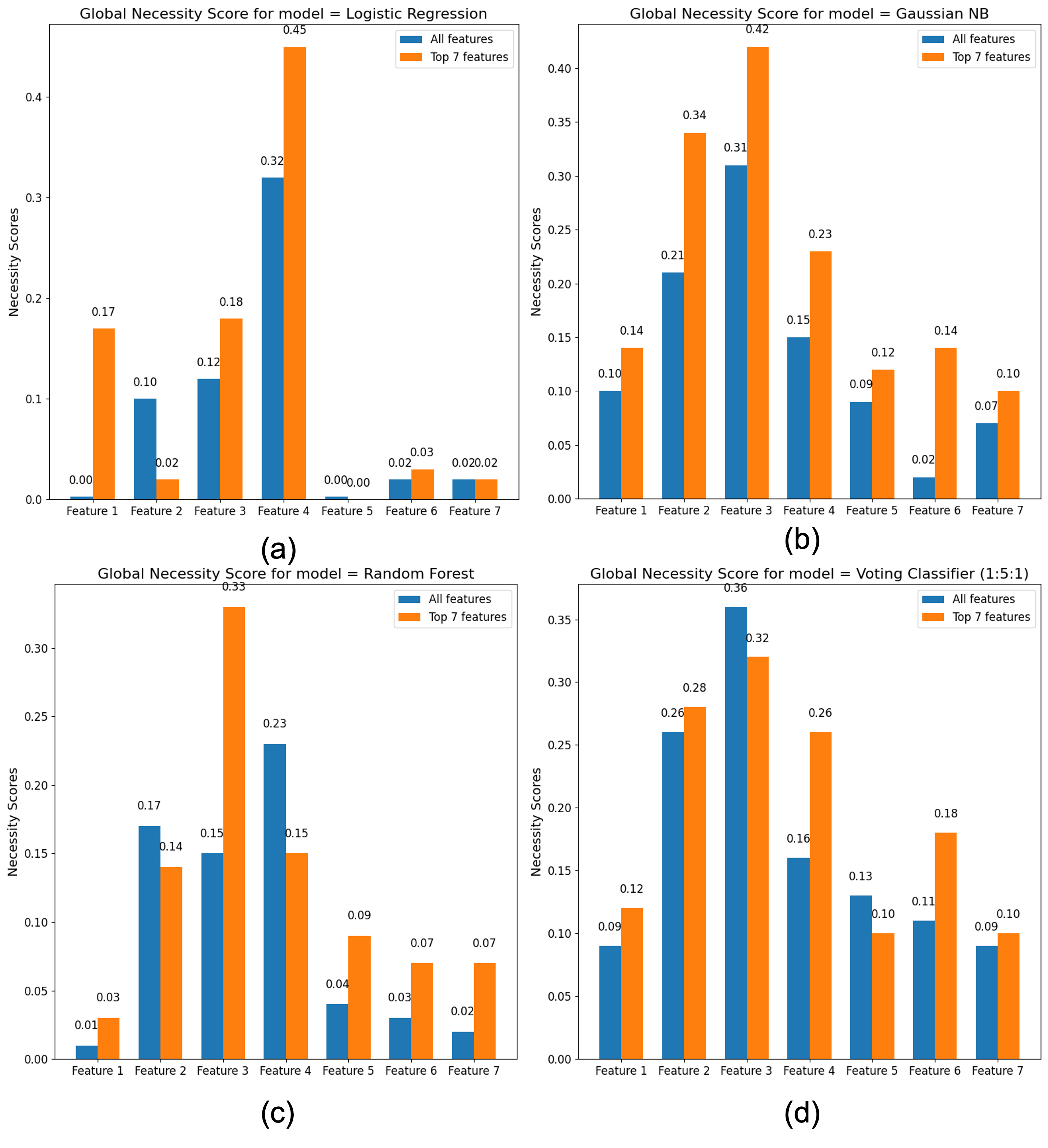} 

  \caption{Global Necessity Score for model (a) Logistic Regression, (b) Gaussian NB, (c) Random Forest and (d) Voting Classifier (1:5:1), for two different settings: ``blue" when model is trained on all features and ``orange" when model is trained on only the top 7 features.}  \label{fig:nec-global}

\end{figure*}

\begin{figure*} 
  \centering
  \includegraphics[width=\textwidth,height=0.55\textheight]{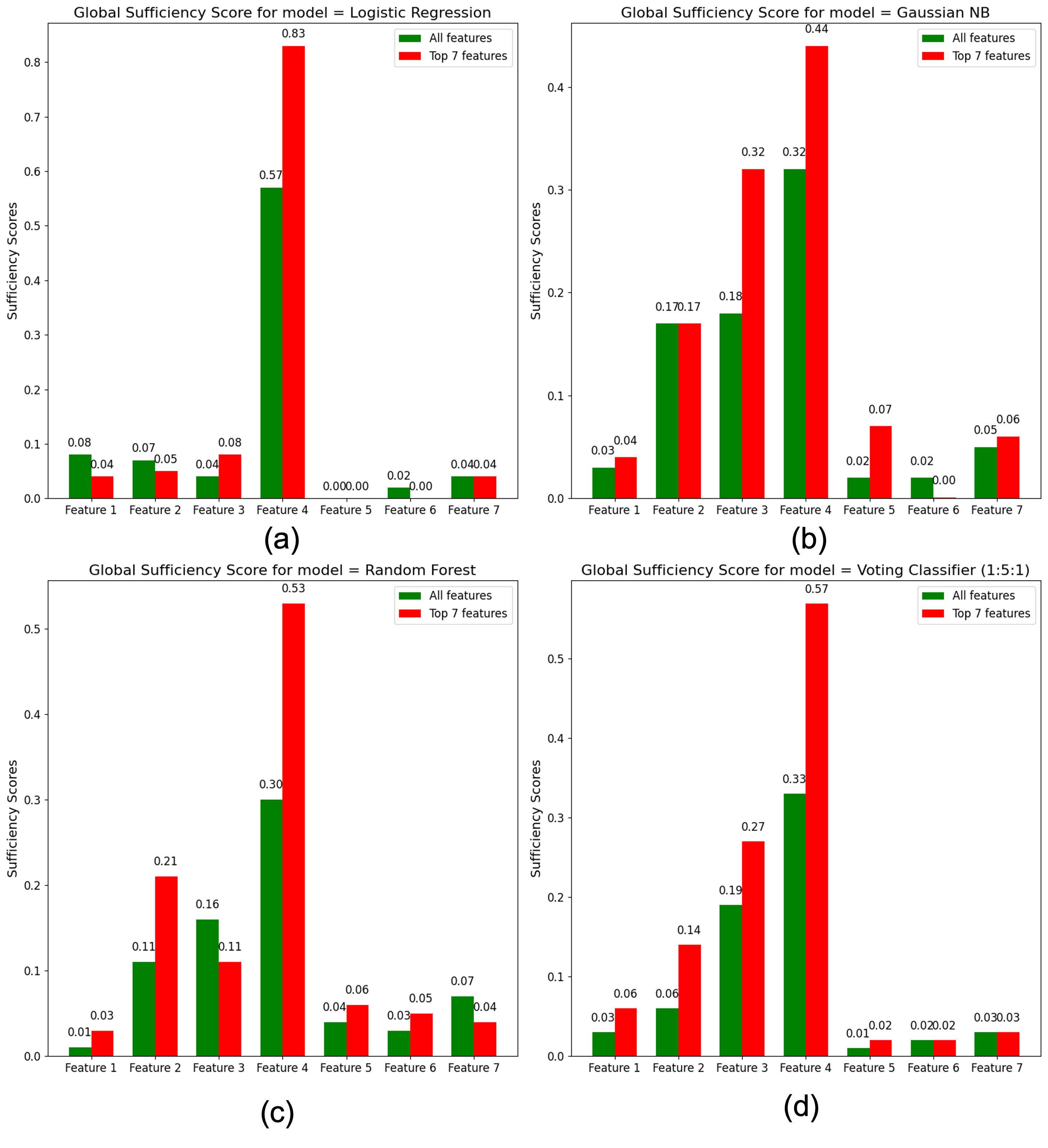} 

  \caption{Global Sufficiency Score for model (a) Logistic Regression, (b) Gaussian NB, (c) Random Forest and (d) Voting Classifier (1:5:1), for two different settings: ``green" when model is trained on all features and ``red" when model is trained on only the top 7 features.}  \label{fig:suf-global}

\end{figure*}

\begin{figure*} 
  \centering
  \includegraphics[width=\textwidth,height=0.55\textheight]{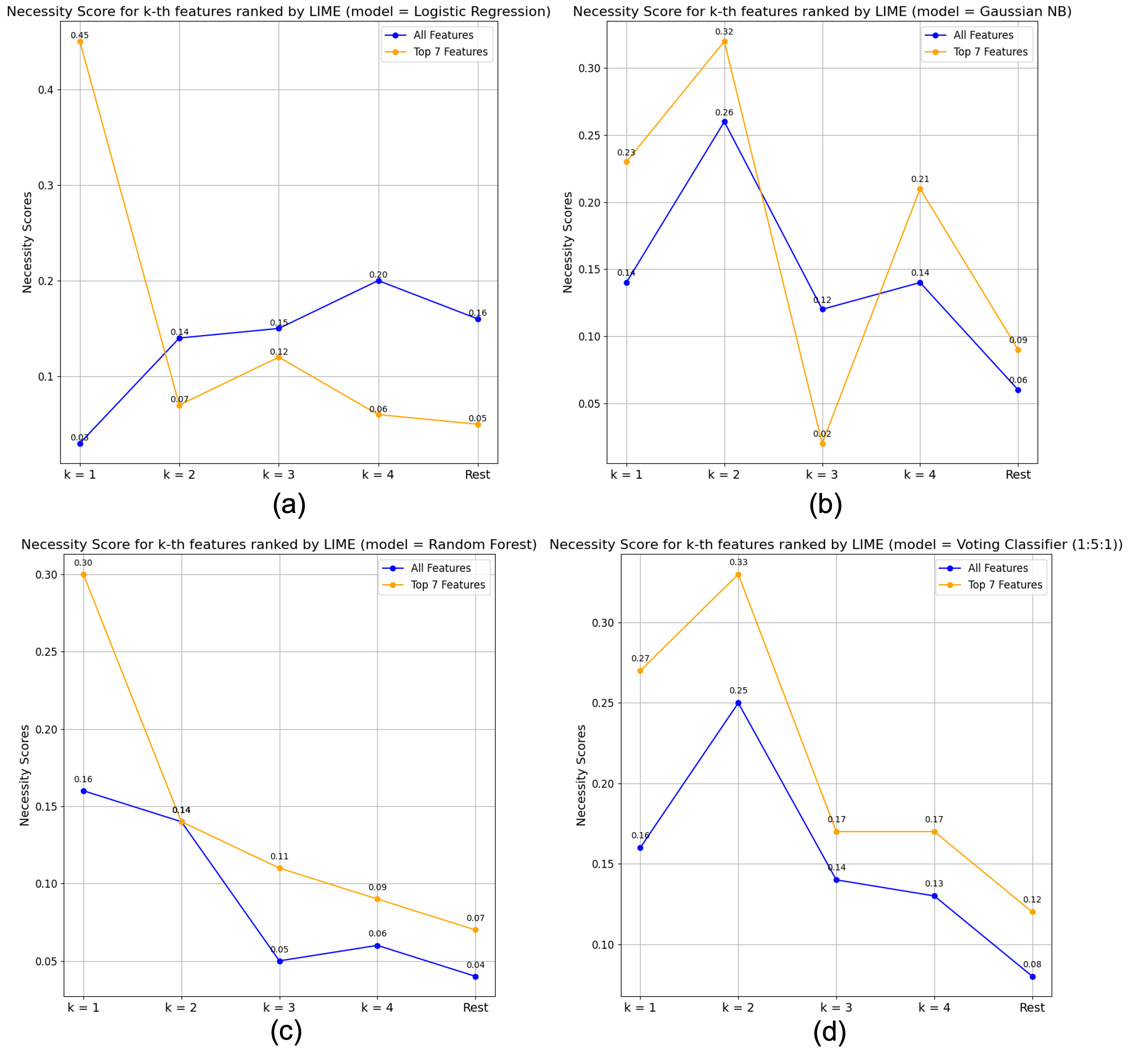} 

  \caption{Necessity Analysis of LIME explanations of predictions made by a (a) Logistic Regression model, a (b) Gaussian Naive Bayes model, a (c) Random Forest Classifier and a (d) Voting Classifier (1:5:1) for two different settings: ``blue" when model is trained on all features and ``orange" when model is trained on only the top 7 features. The necessity scores of the $\#2$ ranked feature in both (b) and (d) are higher than rank $\#1$ thus disagreeing with our hypothesis and failing to be an robust explanation. The LIME explanations conducted on the fitted Random Forest model (c) is observed to be the most robust of the four sets of experiments that have been conducted. } \label{fig:nec-LIME}

\end{figure*}

\begin{figure*} 
  \centering
  \includegraphics[width=\textwidth,height=0.55\textheight]{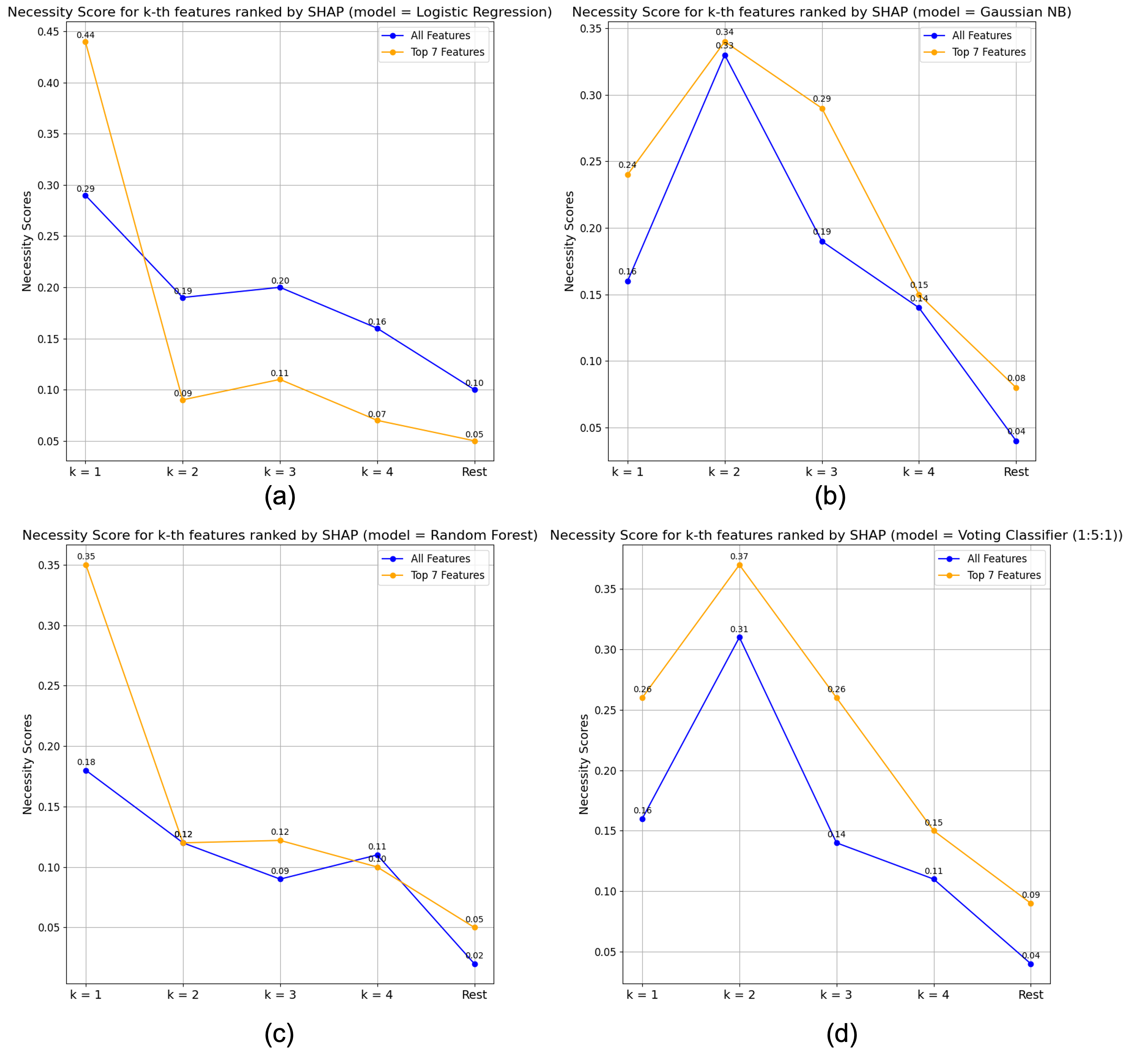} 
  \caption{Necessity Analysis of SHAP explanations of predictions made by a (a) Logistic Regression model, a (b) Gaussian Naive Bayes model, a (c) Random Forest Classifier and a (d) Voting Classifier (1:5:1) for two different settings: ``blue" when model is trained on all features and ``orange" when model is trained on only the top 7 features. The necessity scores of the $\#2$ ranked feature in both (b) and (d) are higher than rank $\#1$ thus disagreeing with our hypothesis and failing to be an robust explanation. The SHAP explanations conducted on the fitted Logistic Regression (a) and Random Forest models (c) are observed to be the robust ones among the four sets of experiments that have been conducted.}  \label{fig:nec-SHAP}
\end{figure*}

\begin{figure*} 
  \centering
  \includegraphics[width=\textwidth,height=0.55\textheight]{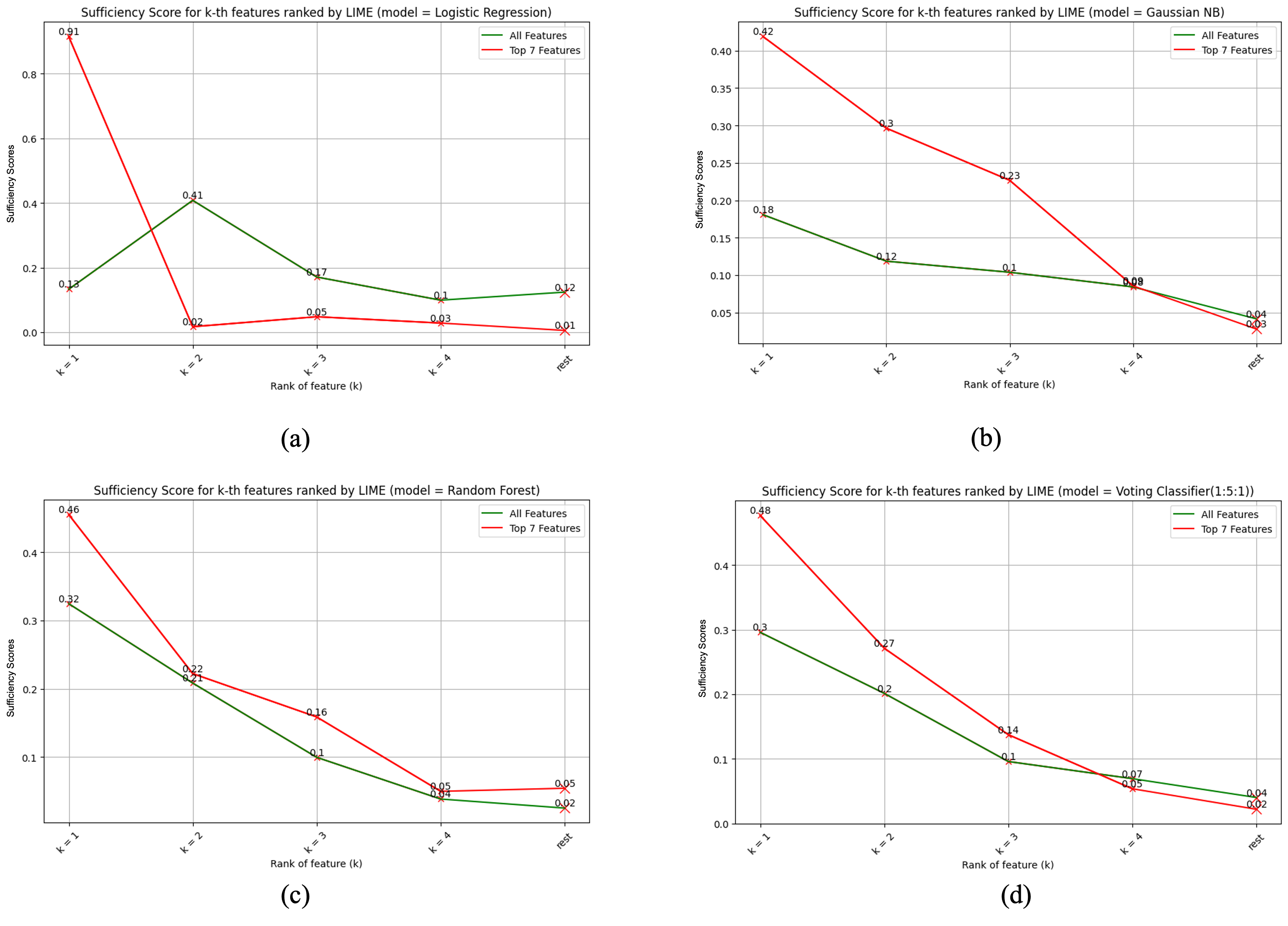} 

  \caption{Sufficiency Analysis of LIME explanations of predictions made by a (a) Logistic Regression model, a (b) Gaussian Naive Bayes model, a (c) Random Forest Classifier and a (d) Voting Classifier (1:5:1) for two different settings: ``green" when model is trained on all features and ``red" when model is trained on only the top 7 features.}  \label{fig:suf-LIME}

\end{figure*}

\begin{figure*} 
  \centering
  \includegraphics[width=\textwidth,height=0.55\textheight]{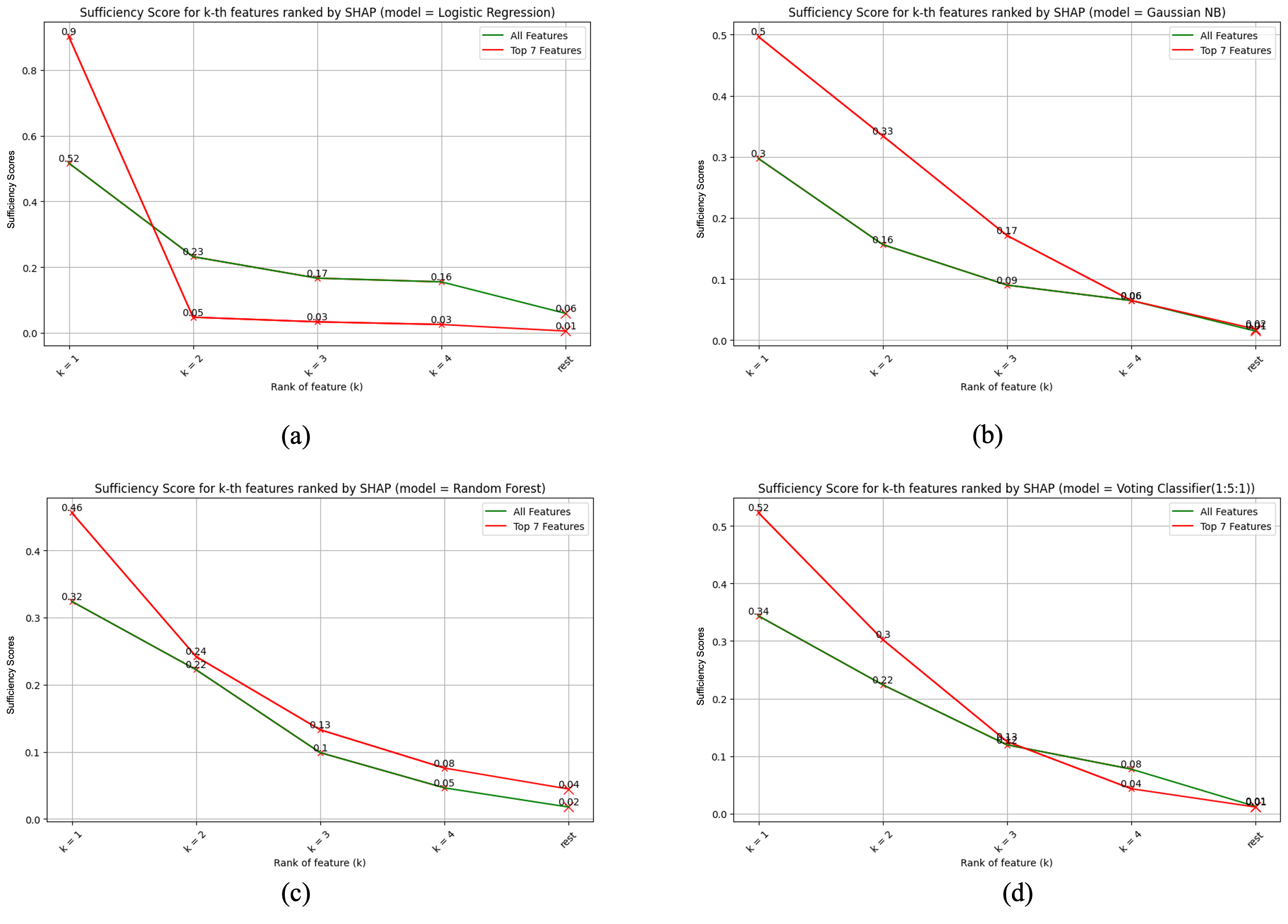} 

  \caption{Sufficiency Analysis of SHAP explanations of predictions made by a (a) Logistic Regression model, a (b) Gaussian Naive Bayes model, a (c) Random Forest Classifier and a (d) Voting Classifier (1:5:1) for two different settings: ``green" when model is trained on all features and ``red" when model is trained on only the top 7 features. The evaluation of the SHAP explanations of all the fitted models are observed to be quite robust to our sufficiency test proving that the feature rankings generated by SHAP correspond to how sufficient a feature is in generating the outcome.}  \label{fig:suf-SHAP}

\end{figure*}









\end{document}